\documentclass[runningheads]{llncs}

\usepackage[T1]{fontenc}
\usepackage{graphicx}
\usepackage{verbatim}  
\usepackage{url}       
\usepackage{xcolor}
\usepackage{listings}
\usepackage{subcaption} 
\usepackage{amsmath}
\usepackage[most]{tcolorbox} 
\usepackage{enumitem}        
\usepackage{xcolor}          
\usepackage{float}
\usepackage{algorithm}
\usepackage{algorithmicx}
\usepackage[noend]{algpseudocode}

\begin{document}

\title{M2-PALE: A Framework for Explaining Multi-Agent MCTS--Minimax Hybrids via Process Mining and LLMs}

\author{
Yiyu Qian\inst{1}\orcidID{0009-0008-8866-7761} \and
Liyuan Zhao\inst{2}\orcidID{0009-0001-3588-7698} \and
Tim Miller\inst{3}\orcidID{0000-0003-4908-6063}
}

\authorrunning{Qian, Zhao, and Miller}

\institute{
RMIT University, Melbourne, Victoria 3000, Australia \\
\email{s4218854@student.rmit.edu.au}
\and
University of California, Irvine, Irvine, CA 92617, USA \\
\email{liyuanz6@uci.edu}
\and
University of Queensland, Brisbane, Queensland 4072, Australia \\
\email{timothy.miller@uq.edu.au}
}

\maketitle

\begin{abstract}
Monte-Carlo Tree Search (MCTS) is a fundamental sampling-based search algorithm widely used for online planning in sequential decision-making domains. Despite its success in driving recent advances in artificial intelligence, understanding the behavior of MCTS agents remains a challenge for both developers and users. This difficulty stems from the complex search trees produced through the simulation of numerous future states and their intricate relationships. A known weakness of standard MCTS is its reliance on highly selective tree construction, which may lead to the omission of crucial moves and a vulnerability to tactical traps. To resolve this, we incorporate shallow, full-width Minimax search into the rollout phase of multi-agent MCTS to enhance strategic depth. Furthermore, to demystify the resulting decision-making logic, we introduce \textsf{M2-PALE} (MCTS--Minimax Process-Aided Linguistic Explanations). This framework employs process mining techniques, specifically the Alpha Miner, iDHM, and Inductive Miner algorithms, to extract underlying behavioral workflows from agent execution traces. These process models are then synthesized by LLMs to generate human-readable causal and distal explanations. We demonstrate the efficacy of our approach in a small-scale checkers environment, establishing a scalable foundation for interpreting hybrid agents in increasingly complex strategic domains.
\end{abstract}

\section{Introduction}

Artificial intelligence (AI) algorithms have demonstrated remarkable efficacy in managing complex tasks across diverse domains~\cite{khan2024leveraging}. However, as these models grow in sophistication, their lack of transparency makes it increasingly difficult to discern how specific outputs are generated~\cite{arrieta2020explainable}. This opacity is commonly termed the ``black-box problem''~\cite{bhatt2020explainable}, signifying that the internal logic of the algorithm remains hidden from both users and stakeholders~\cite{miller2021contrastive}. To mitigate these challenges, Explainable AI (XAI) has emerged as a critical field, aiming to foster user trust, ensure systemic accountability, and promote the ethical and responsible deployment of AI technologies~\cite{bilal2025llms}.

Explainable Reinforcement Learning (XRL) has emerged as a specialized subfield of XAI, dedicated to illuminating the decision-making processes of reinforcement learning (RL) agents. By providing these insights, XRL enables researchers, practitioners, and end-users to effectively understand, validate, and refine learned policies~\cite{cheng2025survey}. A prominent method within this domain is Monte-Carlo Tree Search (MCTS), a model-based planning algorithm utilized for online decision-making in complex sequential environments~\cite{browne2012survey}. However, interpreting the behavior of MCTS agents remains a significant challenge, especially for users without technical background~\cite{an2024enabling}.

Existing research has explored several trajectories to enhance the transparency of MCTS. Initial efforts focused on structural simplification and information-theoretic approaches to reduce the complexity of search trees, making them more amenable to human inspection~\cite{bustin2024structure}. Other approaches have integrated formal logic, such as Computation Tree Logic (CTL), to verify search paths and provide factual or contrastive explanations for sequential planning tasks~\cite{an2024enabling}~\cite{ziyan2025combining}. More recently, the advent of LLMs has enabled the transformation of raw search data into human-readable narratives, bridging the gap between algorithmic complexity and user comprehension~\cite{bilal2025llms}~\cite{gao2410interpretable}.

Despite these advancements, two significant gaps remain. First, while hybrid models, such as the integration of MCTS with Minimax, are increasingly used to bolster tactical robustness in multi-agent or adversarial environments~\cite{baier2014mcts}, their explainability has been largely overlooked. The interplay between MCTS's stochastic exploration and Minimax's deterministic depth-limited search creates a multifaceted decision-making process that existing XRL methods struggle to decompose. Second, most current explanation frameworks are ``state-centric'' or ``path-centric,'' focusing on individual decisions rather than uncovering the broader behavioral patterns or causal workflows inherent in the agent's strategy. This lacks the procedural depth required to explain how an agent navigates long-term dependencies and strategic shifts over time.

To address these challenges, we propose \textsf{M2-PALE} (MCTS--Minimax Process-Aided Linguistic Explanations), a novel framework designed to enhance the interpretability of hybrid decision-making models. \textsf{M2-PALE} incorporates a shallow Minimax search into the rollout phase of multi-agent MCTS to improve strategic depth. It then leverages process mining to extract and interpret the underlying behavioral patterns, ultimately employing LLMs to transform process models into intuitive natural language explanations.
The contributions of this work can be summarized as follows:
\vspace{-4mm}
\begin{itemize}
    \item \textbf{Post-hoc interpretability framework:} We propose a framework that applies post-hoc interpretability to MCTS--Minimax hybrids by leveraging three process discovery algorithms: Alpha Miner, iDHM, and Inductive Miner.
    \item \textbf{LLM-enhanced explanations:} We integrate LLMs to generate causal~\cite{miller2019explanation} and distal explanations~\cite{madumal2020distal} derived directly from the extracted process models, bridging the gap between raw data and human-centric reasoning.
    \item \textbf{Scalability and validation:} Through empirical evaluation in a 3v3 checkers environment, we demonstrate the framework's potential scalability to larger board sizes (6v6, 12v12) and more complex strategic domains.
\end{itemize}

\section{Related Work}


\textbf{Interpretability in MCTS.} Traditional efforts to explain MCTS agents have primarily focused on visualizing or simplifying the search tree. Structural approaches aim to prune less relevant branches or use information-theoretic measures to highlight critical decision nodes~\cite{bustin2024structure}. Recent advancements have introduced formal verification methods to enhance transparency. For instance, An et al.~\cite{an2024enabling} utilized Computation Tree Logic (CTL) to provide contrastive explanations, allowing users to understand why a specific move was chosen over an alternative. However, these methods are often state-centric, focusing on localized decision points. Furthermore, while hybrid models combining MCTS and Minimax are recognized for their tactical superiority in adversarial domains~\cite{baier2014mcts}, their internal decision-making workflows remain under-explored in the XRL literature.

\textbf{Process Mining for AI Transparency.} Process mining has traditionally been used to discover, monitor, and improve real-world business processes. Recently, researchers have begun treating the execution traces of AI agents as event logs to uncover behavioral patterns. Verenich et al.~\cite{verenich2019predicting} demonstrated that converting sequential execution data into process models can provide a ``white-box'' view of a model's performance. Similarly, Gerlach et al.~\cite{gerlach2022inferring} used process mining to evaluate the precision and fitness of next-event predictors. While these studies show the potential of process models to represent complex logic, they rarely extend to the dynamic, stochastic nature of MCTS rollouts or provide high-level linguistic interpretations of the resulting models.

\textbf{LLM-Driven Explanations in XAI.} The integration of LLMs has revolutionized XAI by enabling the translation of complex technical data into intuitive human-readable narratives. Recent frameworks have leveraged LLMs to summarize RL agent policies~\cite{bilal2025llms} or integrated MCTS to elucidate the multi-step reasoning process of LLMs~\cite{gao2410interpretable}. These approaches bridge the gap between algorithmic complexity and user comprehension. However, current LLM-based XRL methods predominantly rely on direct state–action mapping. Such paradigms often yield surface-level explanations that lack causal transparency or a holistic understanding of the agent’s long-term strategic objectives.

\textbf{Distinction and Advantages of M2-PALE.} Compared to the aforementioned works, our proposed \textsf{M2-PALE} framework introduces three distinct advantages:
\begin{enumerate}
    \item \textbf{Hybrid focus:} Unlike previous MCTS-specific or RL-specific interpretability methods, we specifically target MCTS--Minimax hybrids, decomposing the interplay between stochastic exploration and deterministic search.
    \item \textbf{Procedural depth:} By employing process discovery algorithms (Alpha Miner, iDHM, and Inductive Miner), we move beyond state-centric analysis to extract structural workflows that represent the agent's behavioral strategy over time.
    \item \textbf{Causal--distal integration:} While existing LLM methods often describe \emph{what} is happening, \textsf{M2-PALE} uses process models as a grounding mechanism to provide causal and distal explanations, offering deeper insights into the strategic intent behind sequences of moves.
\end{enumerate}

\section{Preliminaries}


\subsection{Process Mining}
Process mining aims to extract functional process models from event logs---records of operational processes consisting of cases and events~\cite{buijs2014quality}. A process model (e.g., Petri-nets) characterizes the behaviors captured in these logs. Recent research has extended process mining to interpret ``black-box'' machine learning models. For instance, Verenich et al.~\cite{verenich2019predicting} utilized process mining for performance indicator prediction via workflow analysis. Similarly, Gerlach et al.~\cite{gerlach2022inferring} applied these techniques to analyze next-event predictors (NEP) through graph-based event logs. However, case generation in complex search spaces remains computationally intensive, which our framework addresses.

\subsection{Quality Dimensions} \label{subsec:quality_dimensions}
\begin{figure}[H]
\centering\includegraphics[width=0.73\linewidth]{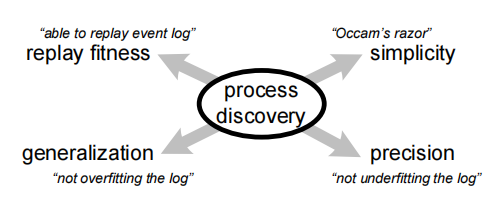}
\caption{Four quality dimensions in process mining~\cite{buijs2014quality}}
\label{fig:quality dimensions}
\end{figure}

To evaluate how effectively a discovered model represents observed behaviors, four quality dimensions are typically measured (Figure~\ref{fig:quality dimensions})~\cite{buijs2014quality}~\cite{van2013mediating}:
\begin{itemize}
\item \textbf{Replay fitness:} Measures the proportion of behaviors in the event log that the process model can accurately reproduce.
\item \textbf{Precision:} Assesses the model's ability to exclude behaviors not observed in the event log, thereby preventing underfitting.
\item \textbf{Generalization:} Estimates the likelihood that the model can describe unseen behaviors generated by the same underlying system, preventing overfitting.
\item \textbf{Simplicity:} Evaluates the model's complexity. Guided by Occam's Razor, a superior model should be as concise as possible while remaining descriptive.
\end{itemize}

While Gerlach et al.~\cite{gerlach2022inferring} use normalized Levenshtein distance and F1-scores to evaluate these dimensions in NEP-generated logs, Buijs et al.~\cite{buijs2014quality} demonstrate that achieving a balance between all four dimensions is critical for robust process discovery.


\subsection{Minimax and MCTS}

Minimax is a recursive decision-making algorithm for adversarial zero-sum games. It constructs a game tree to a specified depth and propagates heuristic evaluation values upward to determine the optimal move, assuming both players act rationally~\cite{strong2011minimax}. Despite its effectiveness, its exponential state-space growth often requires optimizations like $\alpha$--$\beta$ pruning.
Monte Carlo Tree Search (MCTS) approximates action values through random simulations rather than exhaustive depth-first search. MCTS iteratively executes four key steps~\cite{browne2012survey}~\cite{chaslot2008monte}: 1) \textbf{Selection:} Traverses the tree using a selection policy (e.g., UCT) to find the most urgent expandable node~\cite{kocsis2006bandit}; 2) \textbf{Expansion:} Adds one or more child nodes based on available actions; 3) \textbf{Simulation:} Performs a random playout (default policy) from the new node to a terminal state; 4) \textbf{Backpropagation:} Updates the statistics (visit counts and rewards) of all traversed nodes based on the simulation outcome.
These steps continue until a computational budget is reached, at which point the best action (typically the most visited child) is returned.

\begin{algorithm}
	\caption{Minimax Search Algorithm}
	\label{Minimax Search algorithm}
	\begin{algorithmic}[1]
		\Function{Minimax}{$node$, $depth$, $isMaximizingPlayer$}
			\If{terminal state or $depth = 0$}
				\State \Return heuristic value of $node$
			\EndIf
			\If{$isMaximizingPlayer$}
				\State $value \gets -\infty$
				\For{each child $c$ of $node$}
					\State $value \gets \max(value, \Call{Minimax}{c, depth - 1, \mathrm{False}})$
				\EndFor
			\Else
				\State $value \gets \infty$
				\For{each child $c$ of $node$}
					\State $value \gets \min(value, \Call{Minimax}{c, depth - 1, \mathrm{True}})$
				\EndFor
			\EndIf
			\State \Return $value$
		\EndFunction
	\end{algorithmic}
\end{algorithm}

\begin{algorithm}
	\caption{General MCTS approach}
	\label{MCTS Search Algorithm}
	\begin{algorithmic}[1]
		\Function{MCTS}{$s_0$}
			\State create root node $v_0$ with state $s_0$
			\While{within computational budget}
				\State $v_{l} \gets$ \Call{TREEPOLICY}{$v_0$}
				\State $\Delta \gets$ \Call{DEFAULTPOLICY}{$s(v_{l})$}
				\State \Call{BACKUP}{$v_{l}$, $\Delta$}
			\EndWhile
			\State \Return \Call{BESTCHILD}{$v_0$}
		\EndFunction
	\end{algorithmic}
\end{algorithm}

\begin{figure}[!h]
	\centering
	\includegraphics[width=0.82\linewidth]{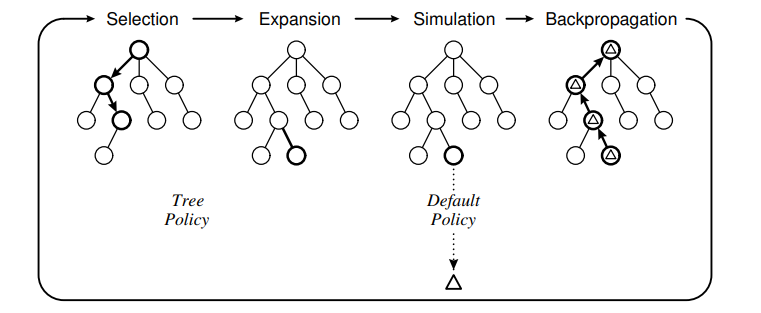}
	\caption{One iteration of the general MCTS approach ~\cite{browne2012survey}}
	\label{fig: One iteration of the general MCTS approach}
\end{figure}


\section{Methodology}


\subsection{Hypothesis}

Our research is guided by a two-fold hypothesis: \textbf{Process model coverage:} Process mining techniques can effectively capture and represent the complete decision-making space explored by MCTS--Minimax hybrids during each turn of a small-scale (e.g., 3v3) checkers game. \textbf{Explainable integration:} Grounded in the extracted process models, LLMs can generate both causal and distal explanations to interpret an agent's strategic actions. Specifically, causal-relationship explanations address fundamental queries regarding action selection and rejection, such as ``Why was this action recommended?'' (Q1) and ``Why was this alternative action dismissed?'' (Q3), by tracing decisions across multiple future game states. Distal explanations address forward-looking queries, such as ``What is the recommended strategy in these potential future scenarios?'' (Q2), based on the procedural patterns identified in the model.

\subsection{General Approach}

As illustrated in Figure~3, our approach aims to discover process models that elucidate the relationships between decisions and predict subsequent strategic shifts. The methodology follows a four-stage pipeline: 1) \textbf{Domain Selection:} Identify a domain with quantifiable state features and action spaces; 2) \textbf{Data Collection and Feature Engineering:} Execute a predefined number of episodes. For each episode, perform feature engineering on the agent's decision-making data and record execution traces; 3) \textbf{Process Discovery and Evaluation:} Aggregate the traces into an event log. Generate process models using diverse algorithms and evaluate their quality via established metrics (fitness, precision, simplicity, and generalization); 4) \textbf{Linguistic Explanation Generation:} Leverage LLMs to transform the structured process models into causal and distal natural language explanations. 

\begin{figure}[!h]
	\centering
	\includegraphics[width=0.66\linewidth]{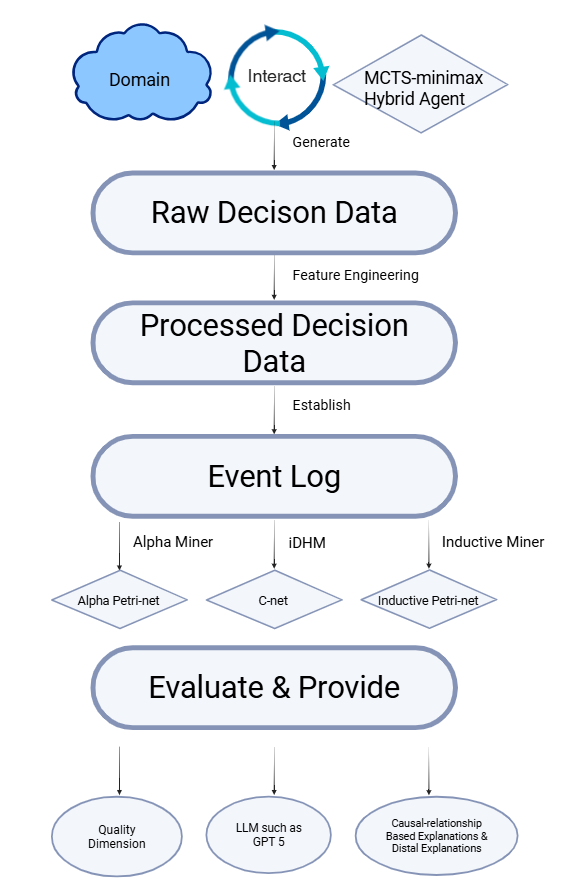}
	\caption{Overview of methodology}
    \vspace{-8mm}
	\label{fig:Overview of methodology}
\end{figure} 

In this study, we apply this approach to the domain of checkers, a non-cooperative board game where two players compete to capture opponent pieces or exhaust their available moves~\cite{samuel1959some}. The implementation details are as follows:
\begin{enumerate}
    \item \textbf{Environment:} We utilize an open-source checkers \footnote{\url{https://github.com/techwithtim/Python-Checkers-AI}}. The state space is defined by coordinates on an $8 \times 8$ grid, and the action space encompasses all valid diagonal movements for each piece.
    \item \textbf{Data Generation:} We execute 100 game episodes. During each turn, the hybrid agent's decisions are recorded and processed through feature engineering to identify the key attributes of the selected moves.
    \item \textbf{Process Discovery:} An event log is synthesized from the 100 episodes. Using the \textit{ProM} framework\footnote{\url{https://promtools.org/}}, we apply three discovery algorithms: \textbf{Alpha Miner}, \textbf{Interactive Discovery of Hybrid Models (iDHM)}, and \textbf{Inductive Miner}. Model performance is rigorously evaluated through conformance checking and performance analysis within the \textit{ProM}.
    \item \textbf{Explanation Synthesis:} We incorporate both the resulting process models and the three specific questions into the prompts for GPT-5. These prompts are designed to elicit natural language insights, including causal explanations for agent behaviors and distal predictions of strategic trajectories.
\end{enumerate}

\begin{lstlisting}[caption={Prompt Template for XAI-driven Process Explanation}, label={lst:prompt_template}]
### ROLE
You are an expert AI Analyst specializing in Process Mining and Explainable AI (XAI). Your task is to interpret formal process models (e.g., Petri-nets, BPMN) to provide rigorous causal and strategic rationales for autonomous agent behavior.

### CONTEXT: PROCESS MODEL INPUT
[Insert Formal Model Representation / State-Transition Logs / Petri Net Markup Here]

### INSTRUCTION: MULTI-LEVEL EXPLANATION GENERATION
Based on the structural patterns and transition probabilities identified in the models, generate detailed technical responses for the following queries:

#### 1. Causal Selection Analysis (Q1)
- **Target:** Why was this specific action recommended?
- **Requirement:** Trace the decision logic through the process flow. Identify the specific control-flow patterns or state-dependencies that justify this transition over others.

#### 2. Contrastive Rejection Analysis (Q3)
- **Target:** Why was this alternative action dismissed?
- **Requirement:** Perform a contrastive explanation. Identify bottlenecks, deadlocks, or sub-optimal terminal states in the process model associated with the rejected path.

#### 3. Distal Strategic Projection (Q2)
- **Target:** What is the recommended strategy for potential future scenarios?
- **Requirement:** Analyze the forward-looking trajectory. Based on the procedural loops and paths, predict the agent's long-term objectives and the convergence of its strategic behavior.

### CONSTRAINTS
- **Groundedness:** All insights must be mathematically or logically derivable from the provided process model.
- **Formality:** Use technical, objective language suitable for a Process Mining research context.
- **Structure:** Explicitly separate immediate causality (Q1, Q3) from distal strategic forecasting (Q2).
\end{lstlisting}

\section{Experiment}

This section details the specific approaches within our methodology and describes the experimental procedures. The primary objective is to evaluate the replay fitness of process models generated by three distinct algorithms: Alpha Miner, iDHM, and Inductive Miner. We leverage these models to provide both causal-relationship-based and distal explanations for the behaviors of a hybrid MCTS–Minimax agent.

\subsection{Domain Setup}

We developed an MCTS--Minimax hybrid agent using Python-based object-oriented programming~\cite{wegner1990concepts}. The agent is encapsulated as a Python class (\texttt{MCTS-Minimax}), where the core phases of MCTS (selection, expansion, simulation, and backpropagation) and the Minimax search algorithm are implemented as class methods.

The game environment is represented by a \texttt{Board} class object containing a two-dimensional array. Each element in the array is an instantiated \texttt{Piece} object defined by four primary attributes:
\begin{enumerate}
    \item \texttt{COLOR} (data type: tuple),
    \item \texttt{PIECE\_ID} (data type: integer),
    \item \texttt{X\_COORDINATE} (data type: integer),
    \item \texttt{Y\_COORDINATE} (data type: integer).
\end{enumerate}
We use RGB $(255, 0, 0)$ for red and $(255, 255, 255)$ for white. The study employs a simplified reward mechanism: 1) Capturing an enemy piece: $+7$ points; 2) Promoting to a crown (king): $+7$ points.
The original checkers game has 12 white pieces and 12 red pieces. We reduce the number of pieces from 12 to 3. Our study is conducted under a 3v3 checkers domain.

\subsection{Minimax and MCTS Integration}

In this study, we perform the strategy of MCTS with Minimax rollouts (MCTS-MR) by integrating shallow Minimax search into the rollout (simulation) stage of multi-agent MCTS~\cite{baier2014mcts}.

\subsection{Data Collection and Feature Engineering}

Each game episode is recorded as a CSV file. We generated 100 episodes for the red agent ($red$ $episode$) and 100 for the white agent ($white$ $episode$). During each turn, the active agent is instantiated as an $MCTS$-$Minimax$ object to determine the optimal move. The red agent's actions are aggregated into a $red$ $eventlog$, and the white agent's actions into a $white$ $eventlog$. In these logs, each unique episode ID is treated as a case, and each discrete action within that episode is treated as an event. Table \ref{tab:episode_data} shows partial red episode and white episode, as well as simplified red event log and simplified white event log. 

Feature engineering is applied to transform raw movement data from the MCTS-Minimax agent into a format suitable for the places and transitions of process models. This process involves selecting, creating, and transforming features to produce high-quality event logs. Our feature engineering workflow consists of four components:

\begin{enumerate}
    \item Spatial Abstraction: Converting exact board coordinates (e.g., $(2, 4) \rightarrow (1, 6)$) into abstract directional representations (e.g., (\texttt{'left'}, \texttt{'up'})).
    \item State Selection: Identifying key features to construct the movement data tuple for the hybrid agent.
    \item Temporal Contextualization: Creating features for the "enemy piece ID" and "enemy movement" from the preceding turn.
    \item Transition Mapping: Selecting specific features to define the transition data tuples within the event log.
\end{enumerate}

For the 3v3 checkers domain, we label each red and white piece uniquely from 1 to 3.

\begin{table*}[htbp]
\centering
\caption{Red and White Episode Data and Event Log. LT=last\_turn, P=piece, C=captured.}
\label{tab:episode_data}

\begin{subtable}{0.50\textwidth}
\centering
\resizebox{\textwidth}{!}{
\begin{tabular}{|c|c|c|c|c|c|}
\hline
\textbf{LT\_id} & \textbf{LT\_movement} & \textbf{P\_id} & \textbf{move} & \textbf{C} & \textbf{reward} \\ \hline
-1 & () & 2 & (`left','down') & [] & 0 \\ \hline
1 & (`right','down') & 3 & (`left','up') & [] & 0 \\ \hline
2 & (`right','down') & 3 & (`left','down') & [] & 0 \\ \hline
1 & (`right','up') & 3 & (`left','down') & [] & 0 \\ \hline
3 & (`right','up') & 3 & (`left','down') & [2] & 14 \\ \hline
1 & (`right','down') & 3 & (`right','up') & [] & 0 \\ \hline
\end{tabular}}
\caption{Partial red episode}
\end{subtable}
\hfill
\begin{subtable}{0.49\textwidth}
\centering
\resizebox{\textwidth}{!}{
\begin{tabular}{|c|c|c|c|c|c|}
\hline
\textbf{LT\_id} & \textbf{LT\_movement} & \textbf{P\_id} & \textbf{move} & \textbf{C} & \textbf{reward} \\ \hline
2 & (`left','down') & 1 & (`right','down') & [] & 0 \\ \hline
3 & (`left','up') & 2 & (`right','down') & [] & 0 \\ \hline
3 & (`left','down') & 1 & (`right','up') & [] & 0 \\ \hline
3 & (`left','down') & 3 & (`right','up') & [] & 0 \\ \hline
3 & (`left','down') & 1 & (`right','down') & [] & 0 \\ \hline
3 & (`right','up') & 1 & (`right','up') & [2] & 7 \\ \hline
\end{tabular}}
\caption{Partial white episode}
\end{subtable}
\vfill
\begin{subtable}{0.49\textwidth}
\centering
\resizebox{\textwidth}{!}{
\begin{tabular}{|c|l|}
\hline
\textbf{task\_id} & \textbf{transition} \\ \hline
1 & ((-1, `()`), (2, `left, up`), 0) \\ \hline
1 & ((3, `right, up`), (3, `left, up`), 0) \\ \hline
1 & ((3, `right, down`), (2, `left, down`), 0) \\ \hline
1 & ((3, `right, up`), (2, `left, up`), 0) \\ \hline
1 & ((1, `right, up`), (3, `left, up`), 7) \\ \hline
2 & ((-1, `()`), (2, `left, down`), 0) \\ \hline
\end{tabular}}
\caption{Simplified red event log}
\end{subtable}
\hfill
\begin{subtable}{0.49\textwidth}
\centering
\resizebox{\textwidth}{!}{
\begin{tabular}{|c|l|}
\hline
\textbf{task\_id} & \textbf{transition} \\ \hline
1 & ((2, `left, up`), (3, `right, up`), 0) \\ \hline
1 & ((3, `left, up`), (3, `right, down`), 0) \\ \hline
1 & ((2, `left, down`), (3, `right, up`), 0) \\ \hline
1 & ((2, `left, up`), (1, `right, up`), 0) \\ \hline
1 & ((3, `left, up`), (1, `right, down`), 0) \\ \hline
2 & ((3, `right, up`), (1, `right, up`), 7) \\ \hline
\end{tabular}}
\caption{Simplified white event log}
\end{subtable}

\end{table*}

\subsection{Trial Design}

We hypothesize that variations in iterations, simulation depth, and Minimax search depth significantly impact the quality of the resulting process models. We executed three trials:
\begin{itemize}
    \item \textbf{Trial 1:} We fix simulation depth at 30 and Minimax search depth at 3. We set the number of iterations to 1000, 2000, and 3000.
    \item \textbf{Trial 2:} We fix the number of iterations at 3000 and Minimax search depth at 3. We set the simulation depth to 10, 20, and 30.
    \item \textbf{Trial 3:} We fix the number of iterations at 3000 and simulation depth at 30. We set the Minimax search depth to 1, 2, and 3.
\end{itemize}

Our baseline assumption is that maximum values for fixed parameters yield the most optimal actions. For each trial, we performed parallel tests using Python’s \texttt{multiprocessing} module~\cite{beazley2010understanding} to ensure true task parallelism. We avoided multi-threading due to the Global Interpreter Lock (GIL)~\cite{beazley2010understanding}, which restricts execution to a single thread at any given time and prevents effective parallelism for CPU-bound tasks. Each individual test within a trial involved the agent playing 100 episodes.

\subsection{Process Model Evaluation}

In this section, we delineate our methodology for evaluating the quality of the generated process models. In each trial, process models for both the red and white agents are synthesized using three distinct algorithms: the Alpha process discovery algorithm, iDHM, and the Inductive Miner algorithm. For each process model, we generate a corresponding replay log utilizing the conformance analysis plug-in provided in \textit{ProM}~\cite{van2005prom}. As established by Van der Aalst et al., conformance analysis compares a process model to an event log of the same process to identify where the real-world process deviates from the modeled behavior~\cite{van2012replaying}. The replay log reports comprehensive global statistics, including the model's replay fitness. A model with high fitness captures most behaviors observed in the event log, whereas a model achieves \textit{perfect fitness} if, and only if, all traces in the log can be replayed by the model from start to finish.

Notably, \textit{ProM} does not currently support conformance analysis for C-nets, which are the process models generated by iDHM. Consequently, replay logs were only obtained for the Petri-nets produced by the Alpha and Inductive Miner algorithms. For every trial, the process models are applied to test our hypothesis. In each test, we focus on three specific values within the global statistics: \textbf{trace fitness}, \textbf{move-log fitness}, and \textbf{move-model fitness}. A process model is classified as a \textit{fitting} model if all three values are perfect (equal to 1), indicating that the model accounts for the vast majority of behaviors in the event log. Conversely, if any of these values is less than 1, the model is categorized as \textit{non-fitting}, suggesting it only represents a minority of the logged behaviors~\cite{ghawi2016process}. 

\begin{figure}[htbp]
    \centering
    
    \begin{subfigure}{0.88\textwidth}
        \centering
        \includegraphics[width=\linewidth]{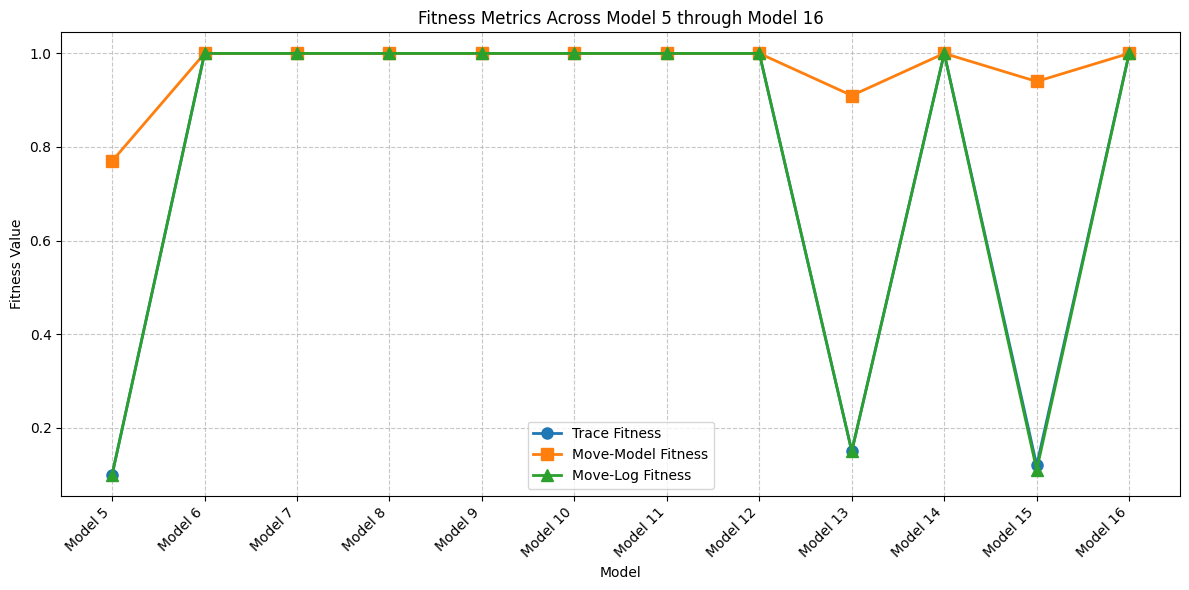}
        \caption{Model 5--16}
        \label{fig:model_5_16}
    \end{subfigure}
    \hfill
    \begin{subfigure}{0.88\textwidth}
        \centering
        \includegraphics[width=\linewidth]{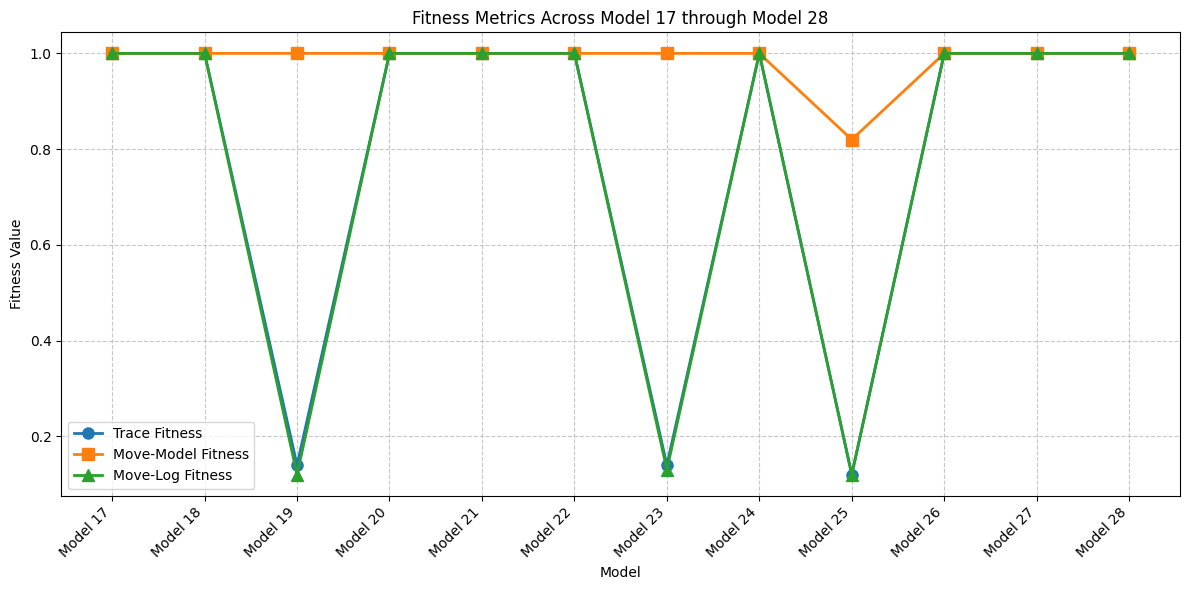}
        \caption{Model 17--28}
        \label{fig:model_17_28}
    \end{subfigure}
    \hfill
    \begin{subfigure}{0.88\textwidth}
        \centering
        \includegraphics[width=\linewidth]{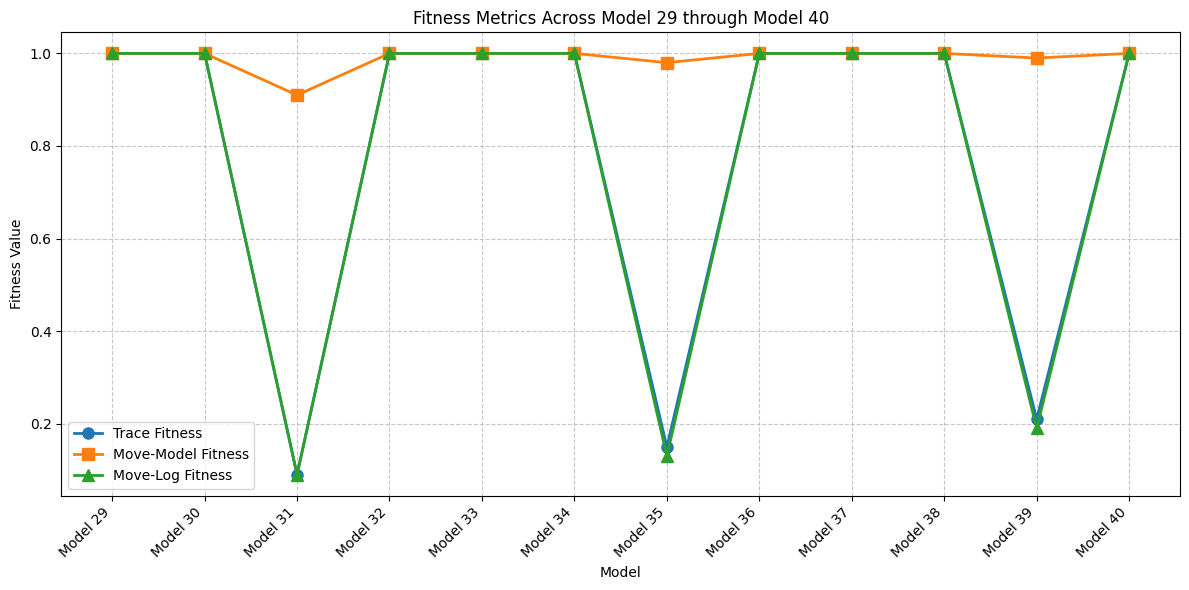}
        \caption{Model 29--40}
        \label{fig:model_29_40}
    \end{subfigure}

    \caption{Visual plots showing fitness metrics across all experimental models (5--40).}
    \label{fig:combined_fitness_plots}
\end{figure}

Table~\ref{tab:petri-net-stats-full} shows global statistics of Petri-nets generated by various algorithms. Table~\ref{tab:combined_global_statistics} shows comparison of global statistics across models 5 to 40. Figure~\ref{fig:combined_fitness_plots} shows visual plots of fitness metrics across all experimental models. Appendix~\ref{Appendix: Evaluate Process Models Based on Trial 1, Trial 2, Trial 3} details how process models are evaluated across Trial~1, Trial~2, and Trial~3.

\vspace{-4mm}

\begin{table*}[!h]
    \centering
    \caption{Global statistics: Petri-net generated by various algorithms (W.A white agent, R.A red agent, I.M inductive miner, A.D alpha discovery)}
    \label{tab:petri-net-stats-full}

    \begin{subtable}{\textwidth}
        \centering
        \resizebox{\textwidth}{!}{
            \begin{tabular}{|l|c|c|c|c|c|c|c|c|c|c|c|c|}
                \hline
                Metric & \multicolumn{4}{c|}{n = 1000} & \multicolumn{4}{c|}{n = 2000} & \multicolumn{4}{c|}{n = 3000} \\
                \cline{2-13}
                & R.A-A.D & R.A-I.M & W.A-A.D & W.A-I.M & R.A-A.D & R.A-I.M & W.A-A.D & W.A-I.M & R.A-A.D & R.A-I.M & W.A-A.D & W.A-I.M \\
                \hline
                Calc. Time (ms)        & 15.74  & 3243.23   & 4.33  & 3738.53   & 4.34   & 2365.69   & 4.50   & 2988.04   & 15.12  & 2906.28   & 12.54  & 3750.52   \\
                Num. States            & 981.20 & 200008.10 & 1.00  & 200019.90 & 1.00   & 200004.50 & 1.00   & 200006.17 & 1189.50 & 200006.70 & 816.90 & 200008.30 \\
                Trace Fitness          & 0.10   & 1.00      & 1.00  & 1.00      & 1.00   & 1.00      & 1.00   & 1.00      & 0.15   & 1.00      & 0.12   & 1.00      \\
                Raw Fitness Cost       & 60.00  & 0.00      & 0.00  & 0.00      & 0.00   & 0.00      & 0.00   & 0.00      & 72.80  & 0.00      & 75.80  & 0.00      \\
                Move-Model Fitness     & 0.77   & 1.00      & 1.00  & 1.00      & 1.00   & 1.00      & 1.00   & 1.00      & 0.91   & 1.00      & 0.94   & 1.00      \\
                Pre-process time (ms)  & 0.40   & 0.34      & 0.33  & 0.59      & 0.51   & 0.46      & 0.30   & 0.37      & 0.52   & 0.48      & 0.43   & 0.39      \\
                Move-Log Fitness       & 0.10   & 1.00      & 1.00  & 1.00      & 1.00   & 1.00      & 1.00   & 1.00      & 0.15   & 1.00      & 0.11   & 1.00      \\
                Trace Length           & 65.90  & 65.90     & 65.20 & 65.20     & 108.33 & 108.33    & 108.33 & 107.83    & 85.30  & 85.30     & 84.90  & 84.90     \\
                Approx. mem. used (kb) & 105.30 & 12007.10  & 62.60 & 11603.90  & 68.50  & 11026.00  & 11026.00 & 10127.00 & 130.90 & 11703.60  & 107.90 & 12266.50  \\
                \hline
            \end{tabular}
        }
        \caption{Fixed Simulation Depth, Fixed Minimax Search Depth, Iteration Times = n}
    \end{subtable}

    \vspace{0.5cm}

    \begin{subtable}{\textwidth}
        \centering
        \resizebox{\textwidth}{!}{
            \begin{tabular}{|l|c|c|c|c|c|c|c|c|c|c|c|c|}
                \hline
                Metric & \multicolumn{4}{c|}{n = 10} & \multicolumn{4}{c|}{n = 20} & \multicolumn{4}{c|}{n = 30} \\
                \cline{2-13}
                & R.A-A.D & R.A-I.M & W.A-A.D & W.A-I.M & R.A-A.D & R.A-I.M & W.A-A.D & W.A-I.M & R.A-A.D & R.A-I.M & W.A-A.D & W.A-I.M \\
                \hline
                Calc. Time (ms)        & 4.07   & 3315.54   & 15.38  & 4072.42   & 4.07   & 3315.54   & 15.38  & 4072.41   & 20.07  & 3430.38   & 5.74   & 3548.60   \\
                Num. States            & 1.00   & 200006.10 & 742.70 & 200010.60 & 1.00   & 200006.10 & 742.70 & 200010.60 & 1333.40 & 200008.50 & 1.00   & 200007.30 \\
                Trace Fitness          & 1.00   & 1.00      & 0.14   & 1.00      & 1.00   & 1.00      & 0.14   & 1.00      & 0.12   & 1.00      & 1.00   & 1.00      \\
                Raw Fitness Cost       & 0.00   & 0.00      & 77.10  & 0.00      & 0.00   & 0.00      & 77.10  & 0.00      & 88.30  & 0.00      & 0.00   & 0.00      \\
                Move-Model Fitness     & 1.00   & 1.00      & 1.00   & 1.00      & 1.00   & 1.00      & 1.00   & 1.00      & 0.82   & 1.00      & 1.00   & 1.00      \\
                Pre-process time (ms)  & 0.69   & 0.38      & 0.30   & 0.52      & 0.69   & 0.38      & 0.30   & 0.52      & 0.29   & 0.48      & 0.32   & 0.57      \\
                Move-Log Fitness       & 1.00   & 1.00      & 0.12   & 1.00      & 1.00   & 1.00      & 0.13   & 1.00      & 0.12   & 1.00      & 1.00   & 1.00      \\
                Trace Length           & 88.90  & 88.90     & 88.60  & 88.60     & 88.90  & 88.90     & 88.60  & 88.60     & 100.30 & 100.30    & 99.90  & 99.90     \\
                Approx. mem. used (kb) & 77.40  & 12607.70  & 105.30 & 11789.60  & 77.40  & 12607.70  & 105.30 & 11789.60  & 161.20 & 12564.30  & 94.50  & 12381.70  \\
                \hline
            \end{tabular}
        }
        \caption{Fixed Iteration Times, Fixed Minimax Search Depth, Simulation Depth = n}
    \end{subtable}

    \vspace{0.5cm}

    \begin{subtable}{\textwidth}
        \centering
        \resizebox{\textwidth}{!}{
            \begin{tabular}{|l|c|c|c|c|c|c|c|c|c|c|c|c|}
                \hline
                Metric & \multicolumn{4}{c|}{n = 1} & \multicolumn{4}{c|}{n = 2} & \multicolumn{4}{c|}{n = 3} \\
                \cline{2-13}
                & R.A-A.D & R.A-I.M & W.A-A.D & W.A-I.M & R.A-A.D & R.A-I.M & W.A-A.D & W.A-I.M & R.A-A.D & R.A-I.M & W.A-A.D & W.A-I.M \\
                \hline
                Calc. Time (ms)        & 3.44   & 3330.23   & 7.48   & 3974.60   & 2.82   & 3566.72   & 14.55  & 5050.51   & 2.79   & 3159.70   & 27.44  & 6170.17   \\
                Num. States            & 1.00   & 200011.00 & 432.30 & 200007.40 & 1.00   & 200008.40 & 844.75 & 200006.85 & 1.00   & 200006.00 & 1296.27 & 200005.33 \\
                Trace Fitness          & 1.00   & 1.00      & 0.09   & 1.00      & 1.00   & 1.00      & 0.15   & 1.00      & 1.00   & 1.00      & 0.21   & 1.00      \\
                Raw Fitness Cost       & 0.00   & 0.00      & 59.40  & 0.00      & 0.00   & 0.00      & 59.30  & 0.00      & 0.00   & 0.00      & 63.33  & 0.00      \\
                Move-Model Fitness     & 1.00   & 1.00      & 0.91   & 1.00      & 1.00   & 1.00      & 0.98   & 1.00      & 1.00   & 1.00      & 0.99   & 1.00      \\
                Pre-process time (ms)  & 0.67   & 0.53      & 0.25   & 0.31      & 0.37   & 0.22      & 0.10   & 0.23      & 0.09   & 0.14      & 0.42   & 0.17      \\
                Move-Log Fitness       & 1.00   & 1.00      & 0.09   & 1.00      & 1.00   & 1.00      & 0.13   & 1.00      & 1.00   & 1.00      & 0.19   & 1.00      \\
                Trace Length           & 64.80  & 64.80     & 64.20  & 64.20     & 72.10  & 72.10     & 67.85  & 67.85     & 101.00 & 101.00    & 78.87  & 78.87     \\
                Approx. mem. used (kb) & 54.80  & 11414.50  & 71.80  & 11856.50  & 51.90  & 12427.20  & 128.05 & 12389.15  & 74.20  & 11796.30  & 214.67 & 12930.40  \\
                \hline
            \end{tabular}
        }
        \caption{Fixed Simulation Depth, Fixed Iteration Times, Minimax Search Depth = n}
    \end{subtable}
\end{table*}

    
    
    
    
    

\begin{table*}[htbp]
	\centering
	\caption{Comparison of Global Statistics across Models 5 to 40}
	\label{tab:combined_global_statistics}

	\begin{subtable}{1\textwidth}
		\centering
		\begin{tabular}{|l|p{2cm}|p{1.2cm}|p{1.2cm}|p{1.2cm}|l|p{1.8cm}|p{1.2cm}|} \hline 
			Model&  Calculation Time (ms)&  Trace Fitness&  Move-Model Fitness&  Move-Log Fitness&  Num. States&  Approx. memory used (kb)& Trace Length\\ \hline 
			Model 5&  15.74&  0.10&  0.77&  0.10&  981.20&  105.30& 65.90\\ \hline 
			Model 6&  3243.23&  1.00&  1.00&  1.00&  200008.10&  12007.10& 65.90\\ \hline 
			Model 7&  4.33&  1.00&  1.00&  1.00&  1.00&  62.60& 65.20\\ \hline 
			Model 8&  3738.53&  1.00&  1.00&  1.00&  200010.90&  11603.90& 65.20\\ \hline 
			Model 9&  4.34&  1.00&  1.00&  1.00&  1.00&  68.50& 108.33\\ \hline 
			Model 10&  2365.69&  1.00&  1.00&  1.00&  200004.50&  11026.00& 108.33\\ \hline 
			Model 11&  4.50&  1.00&  1.00&  1.00&  1.00&  11026.00& 108.33\\ \hline 
			Model 12&  2988.04&  1.00&  1.00&  1.00&  200006.17&  10127.00& 107.83\\ \hline 
			Model 13&  15.12&  0.15&  0.91&  0.15&  1189.50&  130.90& 85.30\\ \hline
			Model 14& 2906.28& 1.00& 1.00& 1.00& 200006.70& 11703.60&85.30\\\hline
			Model 15& 12.54& 0.12& 0.94& 0.11& 816.90& 107.90&84.90\\\hline
			Model 16& 3750.52& 1.00& 1.00& 1.00& 200008.30& 12266.50&84.90\\\hline
		\end{tabular}
		\caption{Global statistics: Model 5 to Model 16}
		\label{tab:fitness-scores-1}
	\end{subtable}

	\vspace{2em} 

	\begin{subtable}{1\textwidth}
		\centering
		\begin{tabular}{|l|p{2cm}|p{1.2cm}|p{1.2cm}|p{1.2cm}|l|p{1.8cm}|l|} \hline 
			Model&  Calculation Time (ms)&  Trace Fitness&  Move-Model Fitness&  Move-Log Fitness&  Num. States&  Approx. memory used (kb)& Trace Length\\ \hline 
			Model 17&  4.07&  1.00&  1.00&  1.00&  1.00&  77.40& 88.90\\ \hline 
			Model 18&  3315.54&  1.00&  1.00&  1.00&  200006.10&  12607.70& 88.90\\ \hline 
			Model 19&  15.38&  0.14&  1.00&  0.12&  742.70&  105.30& 88.60\\ \hline 
			Model 20&  4072.42&  1.00&  1.00&  1.00&  200010.60&  11789.60& 88.60\\ \hline 
			Model 21&  4.07&  1.00&  1.00&  1.00&  1.00&  77.40& 88.90\\ \hline 
			Model 22&  3315.54&  1.00&  1.00&  1.00&  200006.10&  12607.70& 88.90\\ \hline 
			Model 23&  15.38&  0.14&  1.00&  0.13&  742.70&  105.30& 88.60\\ \hline 
			Model 24&  4072.41&  1.00&  1.00&  1.00&  200010.60&  11789.60& 88.60\\ \hline 
			Model 25&  20.07&  0.12&  0.82&  0.12&  1333.40&  161.20& 100.30\\ \hline
			Model 26& 3430.38& 1.00& 1.00& 1.00& 200008.50& 12564.30&100.30\\\hline
			Model 27& 5.74& 1.00& 1.00& 1.00& 1.00& 94.50&99.90\\\hline
			Model 28& 3548.60& 1.00& 1.00& 1.00& 200007.30& 12381.70&99.90\\\hline
		\end{tabular}
		\caption{Global statistics: Model 17 to Model 28}
		\label{tab:fitness-scores-2}
	\end{subtable}

	\vspace{2em}

	\begin{subtable}{1\textwidth}
		\centering
		\begin{tabular}{|l|p{2cm}|p{1.2cm}|p{1.2cm}|p{1.2cm}|c|p{1.8cm}|c|} \hline 
			Model&  Calculation Time (ms)&  Trace Fitness&  Move-Model Fitness&  Move-Log Fitness&  Num. States&  Approx. memory used (kb)& Trace Length\\ \hline 
			Model 29&  3.44&  1.00&  1.00&  1.00&  1.00&  54.80& 64.80\\ \hline 
			Model 30&  3330.23&  1.00&  1.00&  1.00&  200011.00&  11414.50& 64.80\\ \hline 
			Model 31&  7.48&  0.09&  0.91&  0.09&  432.30&  71.80& 64.20\\ \hline 
			Model 32&  3974.60&  1.00&  1.00&  1.00&  200007.40&  11856.50& 64.20\\ \hline 
			Model 33&  2.82&  1.00&  1.00&  1.00&  1.00&  51.90& 72.10\\ \hline 
			Model 34&  3566.72&  1.00&  1.00&  1.00&  200008.40&  12427.20& 72.10\\ \hline 
			Model 35&  14.55&  0.15&  0.98&  0.13&  844.75&  128.05& 67.85\\ \hline 
			Model 36&  5050.51&  1.00&  1.00&  1.00&  200006.85&  12389.15& 67.85\\ \hline 
			Model 37&  2.79&  1.00&  1.00&  1.00&  1.00&  74.20& 101.00\\ \hline
			Model 38& 3159.70& 1.00& 1.00& 1.00& 200006.00& 11796.30&101.00\\\hline
			Model 39& 27.44& 0.21& 0.99& 0.19& 1296.27& 214.67&78.87\\\hline
			Model 40& 6170.17& 1.00& 1.00& 1.00& 200005.33& 12930.40&78.87\\\hline
		\end{tabular}
		\caption{Global statistics: Model 29 to Model 40}
		\label{tab:fitness-scores-3}
	\end{subtable}
\end{table*}

\section{Discussion}



Validating our hypothesis requires a process model that fully captures the behavioral nuances present in the event log. We employed the Inductive Miner algorithm with an iteration count of 3,000, as the resulting Petri-nets for both agents exhibited perfect fitness. For clarity of exposition, the models shown here are based on a simplified log of ten events, ensuring the underlying logic remains accessible without sacrificing behavioral accuracy.
Discussion on white agent and red agent can be referred to Appendix~\ref{Appendix: Discussion Extended}. 

\section{Limitations}
There are some limitations in the research, as follows.
\begin{itemize}
    \item \textbf{Narrow Evaluation of Quality Dimensions}: This study primarily evaluates process models using \textit{replay fitness}. However, a comprehensive assessment should ideally encompass three additional dimensions: (1) \textit{precision}, which reflects the behavioral repertoire in the model not observed in the event log; (2) \textit{simplicity}, which measures human interpretability; and (3) \textit{generalization}, which estimates the model's ability to describe unseen logs from the same system. Due to technical constraints within the \textit{ProM} framework, directly computing these metrics remains challenging. Unlike simplicity, which can be partially inferred from model structure, precision and generalization cannot be accurately assessed through visual inspection of Petri-nets (places and transitions).

    \item \textbf{Absence of Global Win-Rate Analysis}: In this work, explanations are derived from the local rewards associated with specific actions. While we recommend actions yielding the highest immediate rewards, this approach lacks a global perspective on the overall game outcome. A more robust causal explanation would require a measurable insight into how a specific action directly correlates with the final win percentage, providing a global interpretation of the agent's decision-making strategy.

    \item \textbf{Limited Explainability of the Minimax Component}: Although MCTS and Minimax are integrated into a hybrid algorithm, our analysis focuses on how variable Minimax search depths affect process model quality under fixed MCTS iterations. The intrinsic decision-making logic of the Minimax search embedded within the MCTS simulation stage remains under-explored. Future research is needed to dissect the strategy of the Minimax component to better understand its criteria for selecting "optimal" actions during each simulation round.
\end{itemize}

\section{Conclusion}

This research demonstrates that process mining—specifically through the \textit{Inductive Miner} algorithm—provides superior insights into the decision-making strategies of MCTS-Minimax hybrid agents compared to \textit{Alpha Miner} or \textit{iDHM}. The resulting Petri-nets exhibit significant structural coherence, characterized by well-defined source and sink nodes where transitions converge toward a singular outcome. This topology accurately reflects the terminal states inherent in 3v3 checkers episodes. By integrating these process models with LLMs, we have successfully synthesized causal and distal explanations, effectively translating abstract state transitions into interpretable strategic narratives.

While the findings from Trial 1 (Fixed Depth, Variable Iterations) validate the efficacy of our framework within the 3v3 domain, the question of scalability remains. Future work will extend this methodology to more complex domains, such as 6v6 or standard 12v12 checkers, where expanded action spaces may necessitate the pruning operations detailed in Appendix~\ref{Appendix: MCTS Pruning Operation}. To empirically assess the utility of our framework, we intend to conduct a human-centric study. Participants, assisted by LLM-generated explanations, will compete as the Red agent against the MCTS-Minimax hybrid. By quantifying win rates over 100 episodes, we aim to measure the effectiveness of our explanations in enhancing human strategic performance. Finally, we will investigate the generalizability of our post-hoc inquiry framework across diverse sequential decision-making environments and experimental configurations, including variations in simulation and Minimax search depths.

\begin{credits}
\subsubsection{\ackname}
We thank the RMIT hub members for valuable comments and suggestions. We also thank University of Melbourne Associate Professor Artem Polyvyanyy for supervision in this research.

GitHub repository, \url{https://github.com/qyy752457002/Explainable-AI/}.

\end{credits}


\newpage

\renewcommand{\refname}{References}

\bibliographystyle{splncs04}
\bibliography{reference}

\newpage

\appendix
\section{Evaluate Process Models Based on Trial 1, Trial 2, Trial 3}
\label{Appendix: Evaluate Process Models Based on Trial 1, Trial 2, Trial 3}

\subsubsection{Trial 1: Variable Iteration Times} \label{Trial 1}

\begin{itemize}
    \item \textbf{Iteration Times = 1000:} 
    \begin{itemize}
        \item \textbf{Red Agent:} Figure~\ref{fig: red_iteration1000_iDHM} presents the C-Net generated by the iDHM algorithm, while Figure~\ref{fig: red_iteration1000_inductive} illustrates the Petri-net produced by the Inductive Miner. Models 5 and 6 summarize the global statistics for the Red agent using the Alpha Miner and Inductive Miner, respectively. The Petri-net from the Alpha Miner yielded a trace fitness of 0.10, move-model fitness of 0.77, and move-log fitness of 0.10. Consequently, it is classified as a \textit{non-fitting} model as all metrics fall below the threshold of 1.0. In contrast, the Inductive Miner produced a \textit{perfectly fitting} model with all fitness values reaching 1.0.
        
        \item \textbf{White Agent:} Figure~\ref{fig: white_iteration_1000_iDHM} (C-Net) and Figure~\ref{fig: white_iteration_1000_inductive} (Petri-net) display the results for the White agent. Models 7 and 8 provide the corresponding statistics. Both the Alpha Miner and Inductive Miner achieved perfect scores across all fitness metrics (1.0), indicating \textit{fitting} models.
    \end{itemize}

    \item \textbf{Iteration Times = 2000:} 
    \begin{itemize}
        \item \textbf{Red Agent:} The discovered models are shown in Figure~\ref{fig: red_iteration_2000_iDHM} and Figure~\ref{fig: red_iteration_2000_inductive}. Statistics in Models 9 and 10 reveal that both algorithms produced \textit{fitting} models with perfect fitness values (1.0) for all three evaluation criteria.
        
        \item \textbf{White Agent:} Referencing Figure~\ref{fig: white_iteration_1000_iDHM} and Figure~\ref{fig: white_iteration_1000_inductive}, Models 11 and 12 indicate that both algorithms maintained perfect conformance, resulting in \textit{fitting} models.
    \end{itemize}

    \item \textbf{Iteration Times = 3000:}
    \begin{itemize}
        \item \textbf{Red Agent:} As shown in Figure~\ref{fig: red_iteration_3000_iDHM} and Figure~\ref{fig: red_iteration_3000_inductive}, Models 13 and 14 evaluate the Red agent's process. Both models achieved perfect fitness, qualifying as \textit{fitting} models.
        
        \item \textbf{White Agent:} Results are depicted in Figure~\ref{fig: white_iteration_3000_iDHM} and Figure~\ref{fig: white_iteration_3000_inductive}. In Model 15 (Alpha Miner), the trace fitness (0.12), move-model fitness (0.94), and move-log fitness (0.11) indicate a \textit{non-fitting} model. However, Model 16 (Inductive Miner) remains a \textit{fitting} model with perfect scores.
    \end{itemize}
\end{itemize}

\textbf{Perfect Fitness Models (6--12, 14, 16):} These models demonstrate ideal conformance to the event logs. Maximum fitness values (1.0) across Trace, Move-Model, and Move-Log metrics suggest that these process models accurately capture both the sequential flow of activities and individual transitions within the 3v3 checkers domain.

\textbf{Non-Perfect Fitness Models (5, 13, 15):} Model 5 exhibits the most significant deviation, with minimal scores across all metrics, indicating a failure to represent the actual process logic. Model 13 shows a discrepancy where high Move-Model fitness (0.91) suggests individual transitions are captured, yet low Trace Fitness (0.15) reveals poor representation of the overall activity sequences. Similarly, Model 15 reflects high transition accuracy (0.94) but fails to align with the global execution sequences in the log.

\subsubsection{Trial 2: Fixed Iteration Times, Fixed Minimax Search Depth, Variable Simulation Depth} \label{Trial 2}

\begin{itemize}
    \item \textbf{Simulation Depth = 10:} 
    \begin{itemize}
        \item Both Red (Models 17, 18) and White (Models 19, 20) agents achieved perfect conformance using both algorithms. Figures~\ref{fig: red_simulation_10_iDHM} through~\ref{fig: white_simulation_10_inductive} visualize these \textit{fitting} models.
    \end{itemize}
    
    \item \textbf{Simulation Depth = 20:} 
    \begin{itemize}
        \item \textbf{Red Agent:} Models 21 and 22 show perfect fitness for both algorithms.
        \item \textbf{White Agent:} In Model 23 (Alpha Miner), while move-model fitness reached 1.0, trace fitness (0.14) and move-log fitness (0.12) were significantly lower, classifying it as \textit{non-fitting}. Model 24 (Inductive Miner) maintained perfect fitness.
    \end{itemize}
    
    \item \textbf{Simulation Depth = 30:}
    \begin{itemize}
        \item \textbf{Red Agent:} Model 25 (Alpha Miner) is \textit{non-fitting} with a move-model fitness of 0.82 and trace/move-log fitness of 0.12. Model 26 (Inductive Miner) remains perfectly \textit{fitting}.
        \item \textbf{White Agent:} Models 27 and 28 both show perfect conformance.
    \end{itemize}
\end{itemize}

\textbf{Consistent High Fitness:} The majority of the models (17, 18, 20--22, 24, 26--28) exhibit perfect alignment (1.0) with the experimental data. This consistency indicates that the discovered models are highly robust in representing the agent's decision-making process under varying simulation depths.

\textbf{Performance Outliers:} Significant deviations were observed in Models 19, 23, and 25. In Models 19 and 23, the sharp decline in Move-Log fitness despite perfect Move-Model fitness suggests that while the model structure is theoretically sound, it does not align with the observed execution frequencies in the logs. Model 25 represents a more severe misalignment, with deficiencies in both structural representation and log replay. Interestingly, Trace Fitness remains consistently high at 1.00 across most models, suggesting that the tracing aspect of the model performs well regardless of other fluctuations.

\subsubsection{Trial 3: Fixed Simulation Depth, Fixed Iteration Times, Variable Minimax Search Depth} \label{Trial 3}

\begin{itemize}
    \item \textbf{Minimax Search Depth = 1:} 
    \begin{itemize}
        \item \textbf{Red Agent:} Model 29 (Alpha Miner) is \textit{non-fitting} (Trace: 0.94, Move-Model: 0.09). Model 30 (Inductive Miner) is a perfect \textit{fit}.
        \item \textbf{White Agent:} Models 31 and 32 both show perfect conformance.
    \end{itemize}
    
    \item \textbf{Minimax Search Depth = 2:} 
    \begin{itemize}
        \item \textbf{Red Agent:} Models 33 and 34 both exhibit perfect \textit{fitting}.
        \item \textbf{White Agent:} Model 35 (Alpha Miner) is \textit{non-fitting} (Trace: 0.15, Move-Log: 0.13), whereas Model 36 (Inductive Miner) is perfectly \textit{fitting}.
    \end{itemize}
    
    \item \textbf{Minimax Search Depth = 3:}
    \begin{itemize}
        \item \textbf{Red Agent:} Models 37 and 38 both maintain perfect conformance.
        \item \textbf{White Agent:} Model 39 (Alpha Miner) shows a move-model fitness of 0.99 but low trace (0.21) and move-log (0.19) fitness. Model 40 (Inductive Miner) remains perfectly \textit{fitting}.
    \end{itemize}
\end{itemize}

\textbf{Perfect Model Conformance:} Models 29, 30, 32--34, 36--38, and 40 demonstrate exceptional performance with fitness values of 1.00. This indicates that the models accurately track real-time agent data and that the transition sequences are fully consistent with the observed game logs.

\textbf{Non-fitting Models and Logical Divergence:} Models 31, 35, and 39 stand out as outliers. The significant drops in Trace and Move-Log fitness in Model 31 suggest low accuracy in mapping the model to empirical data. In Model 39, the high Move-Model fitness (0.99) alongside low Trace fitness (0.21) reveals a logical divergence: while individual moves are captured, the aggregate sequences generated by the deeper Minimax search deviate significantly from standard expectations.

\section{Discussion Extended}
\label{Appendix: Discussion Extended}

For red agent in Figure~\ref{fig: red_simplified} , we can see transitions in the first layer are

\begin{itemize}
	\item $((-1, ``0"), (2, (``left", ``down")), 0)$
	\item $((-1, ``0"), (2, (``left", ``up")), 0)$
	\item $((-1, ``0"), (3, (``left", ``down")), 0)$
\end{itemize}

and transitions in the second layer are 

\begin{itemize}
	\item $((3, (``right", ``up")),\allowbreak (1, (``left", ``up")), 7)$
	\item $((1, (``left", ``down")),\allowbreak (2, (``left", ``up")), 7)$
	\item $((3, (``right", ``down")),\allowbreak (2, (``left", ``down")), 0)$
	\item $((1, (``right", ``down")),\allowbreak (3, (``right", ``up")), 0)$
\end{itemize}

and transitions in the third layer are 

\begin{itemize}
	\item $((2, (``left", ``down")),\allowbreak (1, (``left", ``up")), 7)$
	\item $((1, (``left", ``up")),\allowbreak (1, (``left", ``down")), 0)$
\end{itemize}

For white agent in Figure~\ref{fig: white_simplified}, we can see the transition in the first layer is

\begin{itemize}
	\item $((2, (``left", ``down")),\allowbreak (2, (``right", ``up")), 0)$
\end{itemize}

and transitions in the second layer are 

\begin{itemize}
	\item $((2, (``left", ``up")),\allowbreak (3, (``right", ``down")), 0)$
	\item $((2, (``right", ``up")),\allowbreak (3, (``left", ``down")), 0)$
	\item $((3, (``left", ``down")),\allowbreak (1, (``right", ``up")), 0)$
	\item $((3, (``left", ``down")),\allowbreak (2, (``right", ``up")), 7)$
\end{itemize}

\begin{tcolorbox}[
    colback=gray!5, 
    colframe=black!75, 
    title={GPT-5 Generated Strategic Interpretation Report},
    fonttitle=\bfseries,
    arc=5pt,
    outer arc=5pt
]

\small

\subsection*{1. Red Agent Strategic Analysis (Ref: Figure~\ref{fig: red_simplified})}
The following insights are derived from the hierarchical transition layers of the Red agent's Petri-net:

\begin{description}[style=multiline, leftmargin=3cm, font=\bfseries]
    \item[Causal Selection (Q1)] The recommendation to select \textbf{Piece 1 (left, up)} or \textbf{Piece 2 (left, up)} in the second layer is driven by immediate reward optimization. These transitions yield \textbf{7 reward points}, correlating directly with high-value outcomes such as capturing an enemy piece or achieving a crown.
    
    \item[Causal Rejection (Q3)] Alternative actions, specifically $(2, (\text{``left", ``down"}))$ and $(3, (\text{``right", ``up"}))$, are dismissed. The model identifies these as non-productive transitions ($0$ points) that fail to improve the agent's material standing in the current tactical window.
    
    \item[Distal Strategy (Q2)] The model demonstrates "look-ahead" reasoning for future scenarios. While choosing \textbf{Piece 2 (left, down)} in response to a right-down move by the opponent yields an immediate $0$ reward, it is classified as a \textbf{provident action}. As evidenced in the third layer's transition $((2, (``left", ``down")), (1, (``left", ``up")), 7)$, this path is a necessary precursor to securing 7 points in subsequent turns.
\end{description}


\subsection*{2. White Agent Strategic Analysis (Ref: Figure~\ref{fig: white_simplified})}
The interpretation of the White agent's procedural patterns is summarized as follows:

\begin{description}[style=multiline, leftmargin=3cm, font=\bfseries]
    \item[Causal Selection (Q1)] Upon the Red agent moving Piece 3 (left, down), the system recommends \textbf{White Piece 2 (right, up)}. This selection is justified by its potential to trigger a capture or crowning event, identified as a 7-point reward path in the second transition layer.
    
    \item[Causal Rejection (Q3)] Transitions such as $(3, (\text{``right", ``down"}))$ and $(1, (\text{``right", ``up"}))$ are rejected. The Petri-net classifies these as suboptimal branches that do not result in immediate point acquisition ($0$ points).
    
    \item[Distal Strategy (Q2)] In future scenarios where Red advances Piece 3, the agent recommends selecting \textbf{White Piece 2 (left, up)} as a provident move to lock in 7 reward points. For complex branches (e.g., Red moving Piece 2), the system suggests selecting \textbf{Piece 3}. To confirm optimality, the model advises a distal search into the **third and fourth layers** to identify which trajectory yields the earliest reward emergence.
\end{description}

\end{tcolorbox}

\section{MCTS Pruning Operation}\label{Appendix: MCTS Pruning Operation}

\begin{figure}[H]
	\centering
	\includegraphics[width=0.82\linewidth]{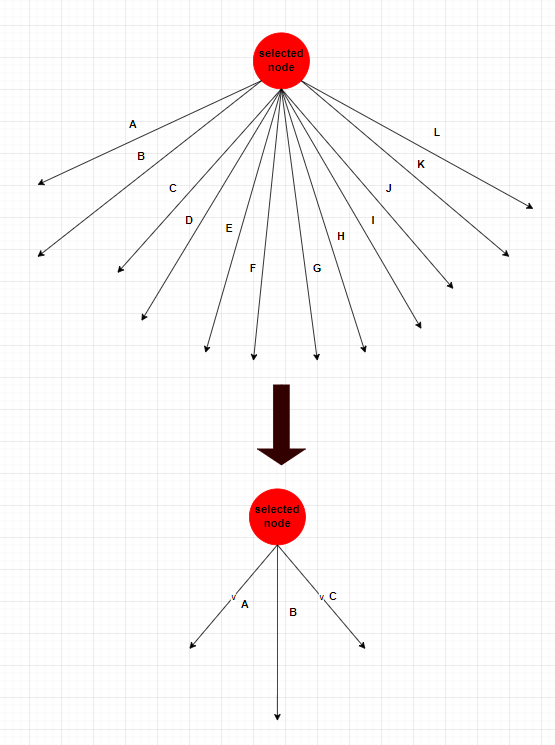}
	\caption{Pruning operation}
	\label{Pruning operation}
\end{figure}

\begin{figure}[H]
	\centering
	\includegraphics[width=1.3\linewidth]{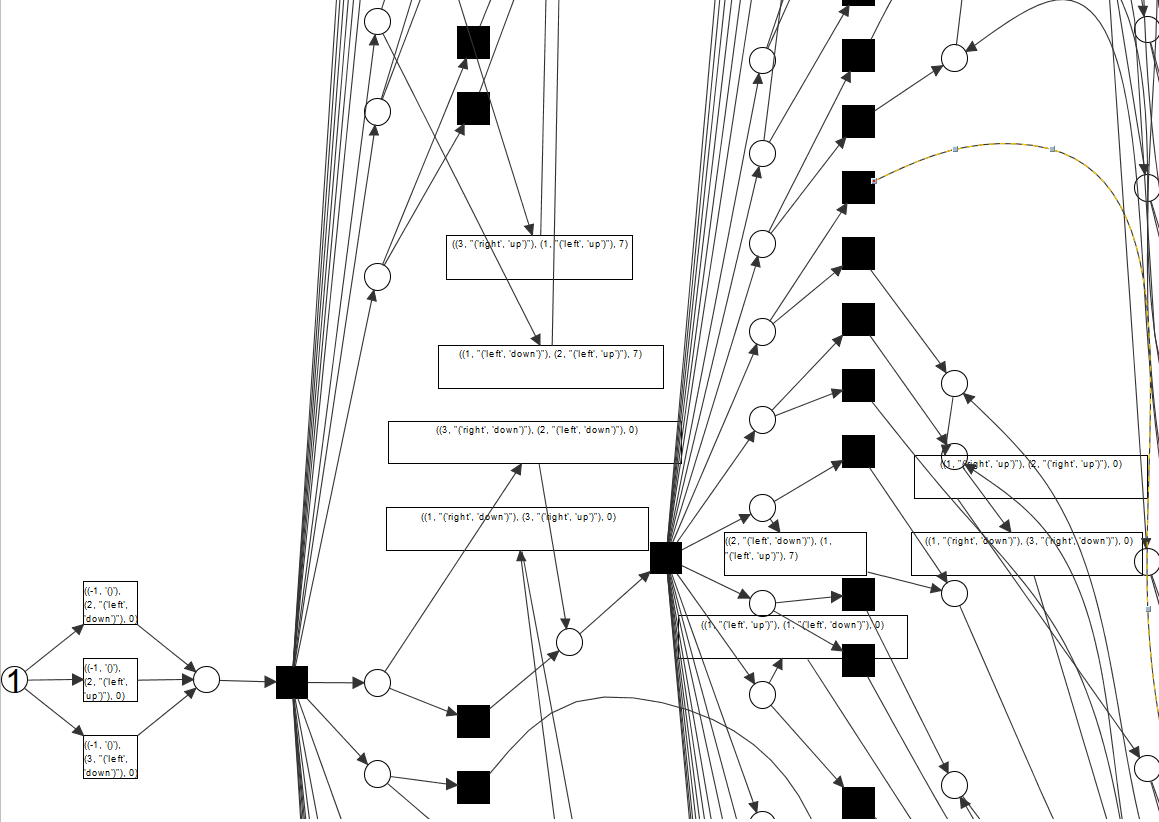}
	\caption{Simplified red agent Petri-net (10 episodes) generated by inductive miner algorithm (Fixed Simulation Depth, Fixed Minimax Search Depth, Iteration Times = 3000)}
	\label{fig: red_simplified}
\end{figure}

\begin{figure}[!h]
	\centering
	\includegraphics[width=1.3\linewidth]{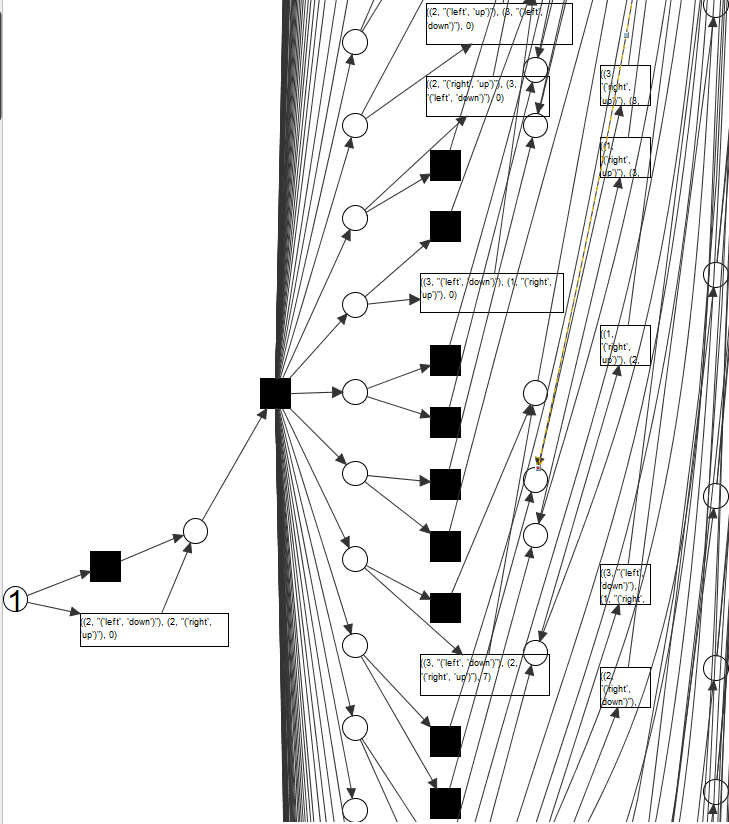}
	\caption{Simplified white agent Petri-net (10 episodes) generated by inductive miner algorithm (Fixed Simulation Depth, Fixed Minimax Search Depth, Iteration Times = 3000)}
	\label{fig: white_simplified}
\end{figure}

\begin{figure}[!h]
	\centering
	\includegraphics[width=1\linewidth]{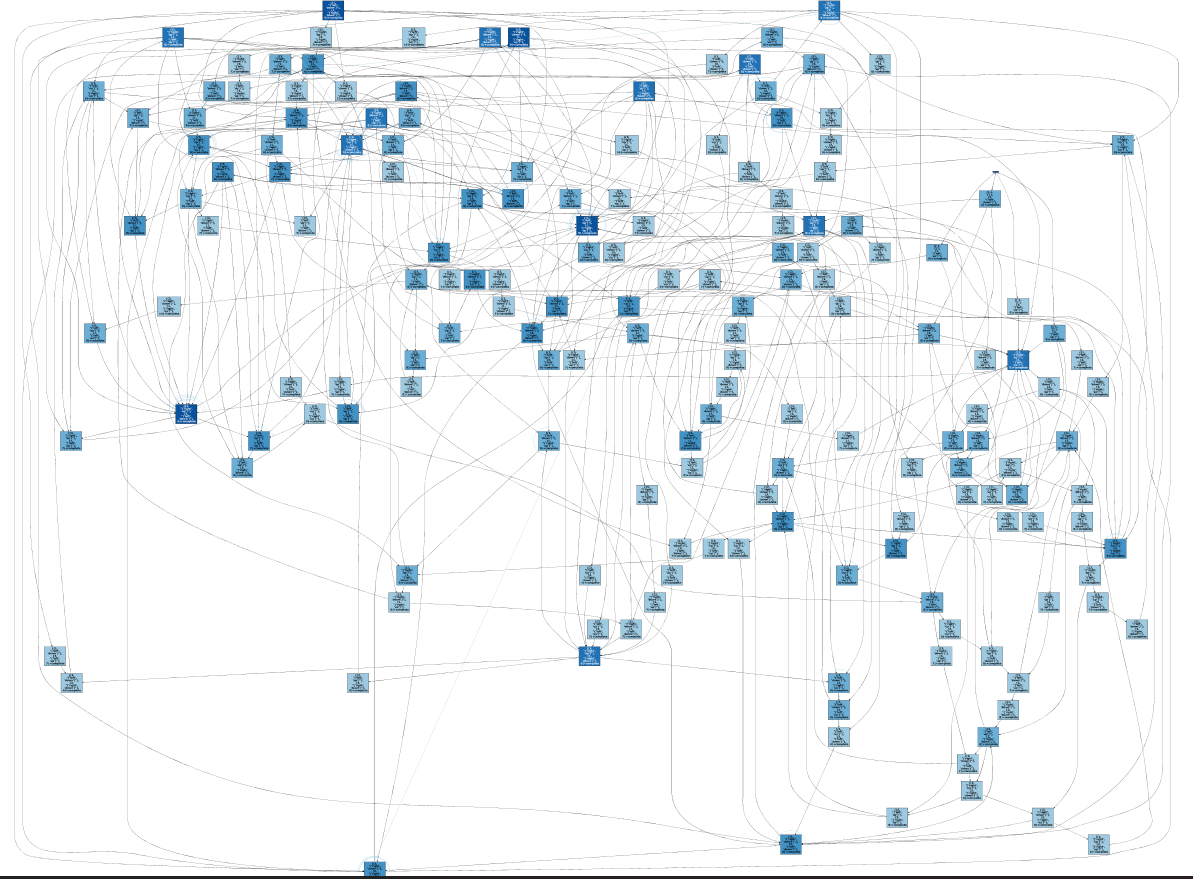}
	\caption{Red agent C-net generated by iDHM (Fixed Simulation Depth, Fixed Minimax Search Depth, Iteration Times = 1000)}
	\label{fig: red_iteration1000_iDHM}
\end{figure}

\begin{figure}[!h]
	\centering
	\includegraphics[width=1\linewidth]{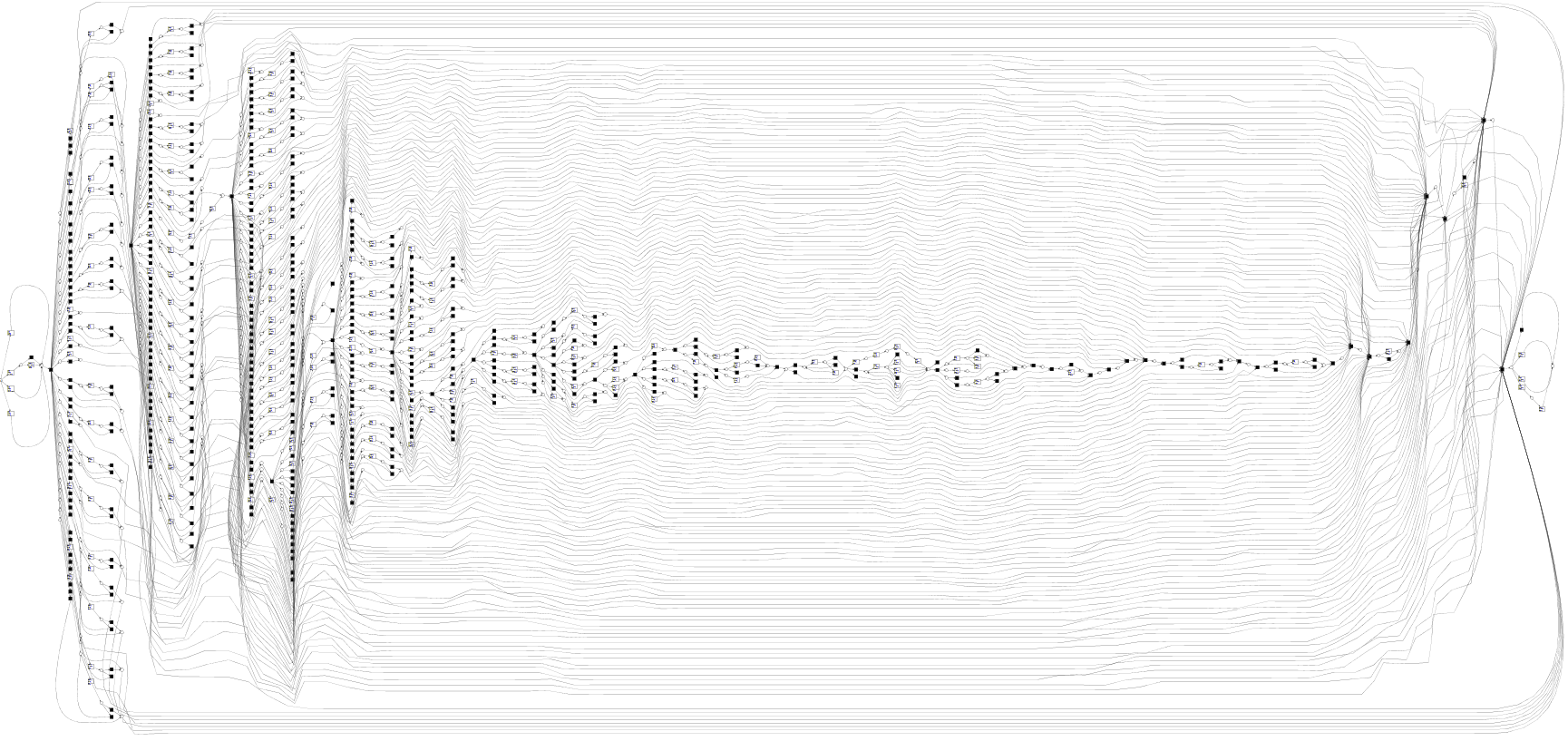}
	\caption{Red agent Petri-net generated by inductive miner algorithm (Fixed Simulation Depth, Fixed Minimax Search Depth, Iteration Times = 1000)}
	\label{fig: red_iteration1000_inductive}
\end{figure}

\begin{figure}[!h]
	\centering
	\includegraphics[width=1\linewidth]{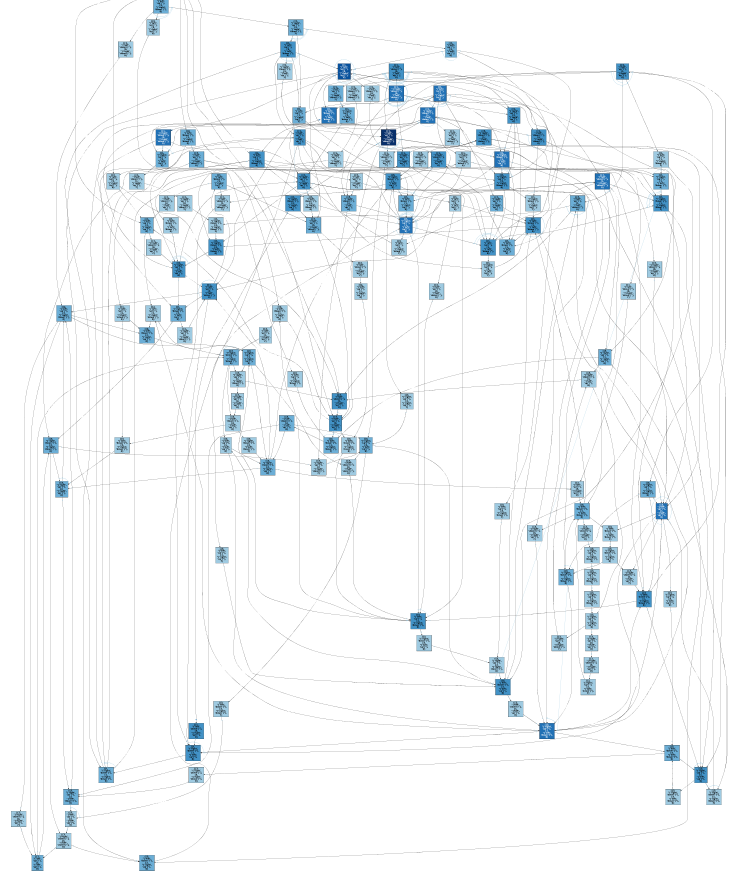}
	\caption{White agent C-net generated by iDHM (Fixed Simulation Depth, Fixed Minimax Search Depth, Iteration Times = 1000)}
	\label{fig: white_iteration_1000_iDHM}
\end{figure}

\begin{figure}[!h]
	\centering
	\includegraphics[width=1\linewidth]{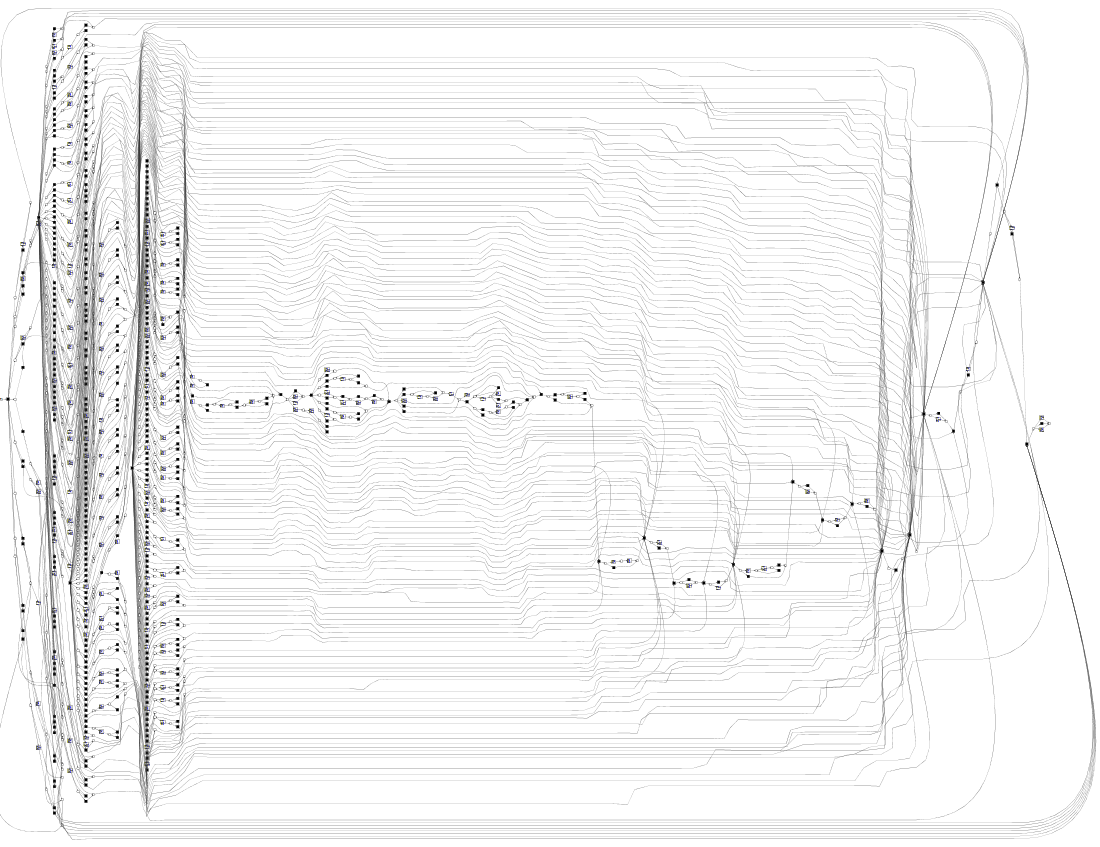}
	\caption{White agent Petri-net generated by inductive miner algorithm (Fixed Simulation Depth, Fixed Minimax Search Depth, Iteration Times = 1000)}
	\label{fig: white_iteration_1000_inductive}
\end{figure}

\begin{figure}[!h]
	\centering
	\includegraphics[width=1\linewidth]{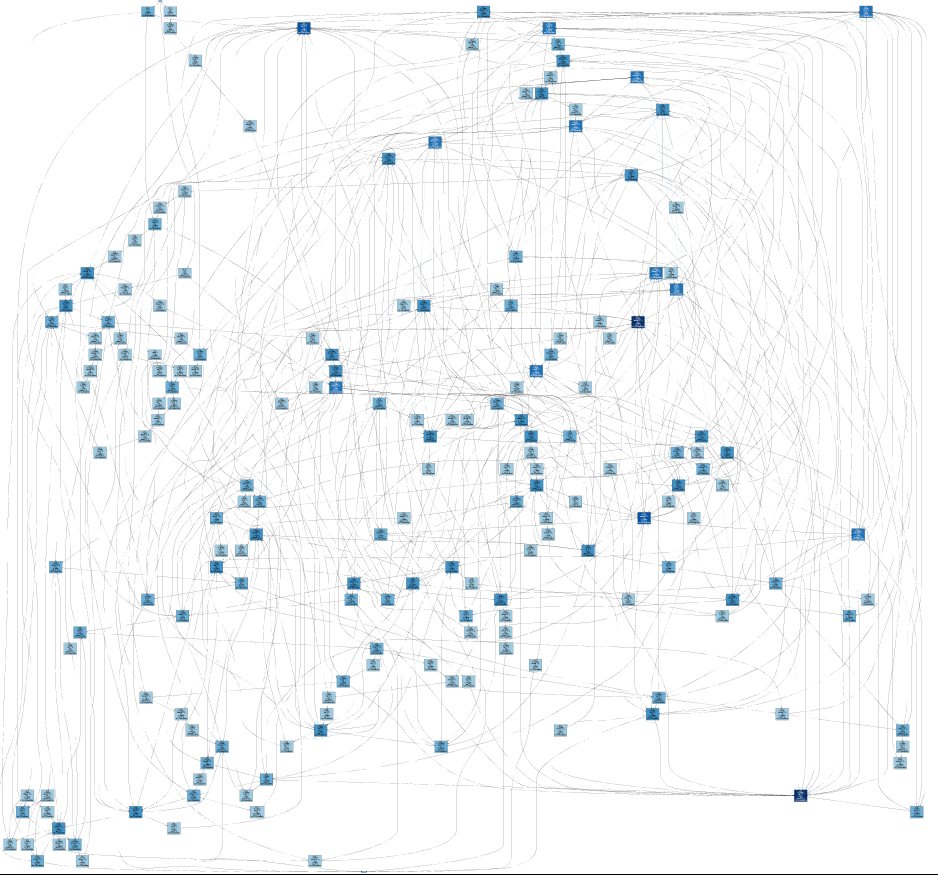}
	\caption{Red agent C-net generated by iDHM (Fixed Simulation Depth, Fixed Minimax Search Depth, Iteration Times = 2000)}
	\label{fig: red_iteration_2000_iDHM}
\end{figure}

\begin{figure}[!h]
	\centering
	\includegraphics[width=1\linewidth]{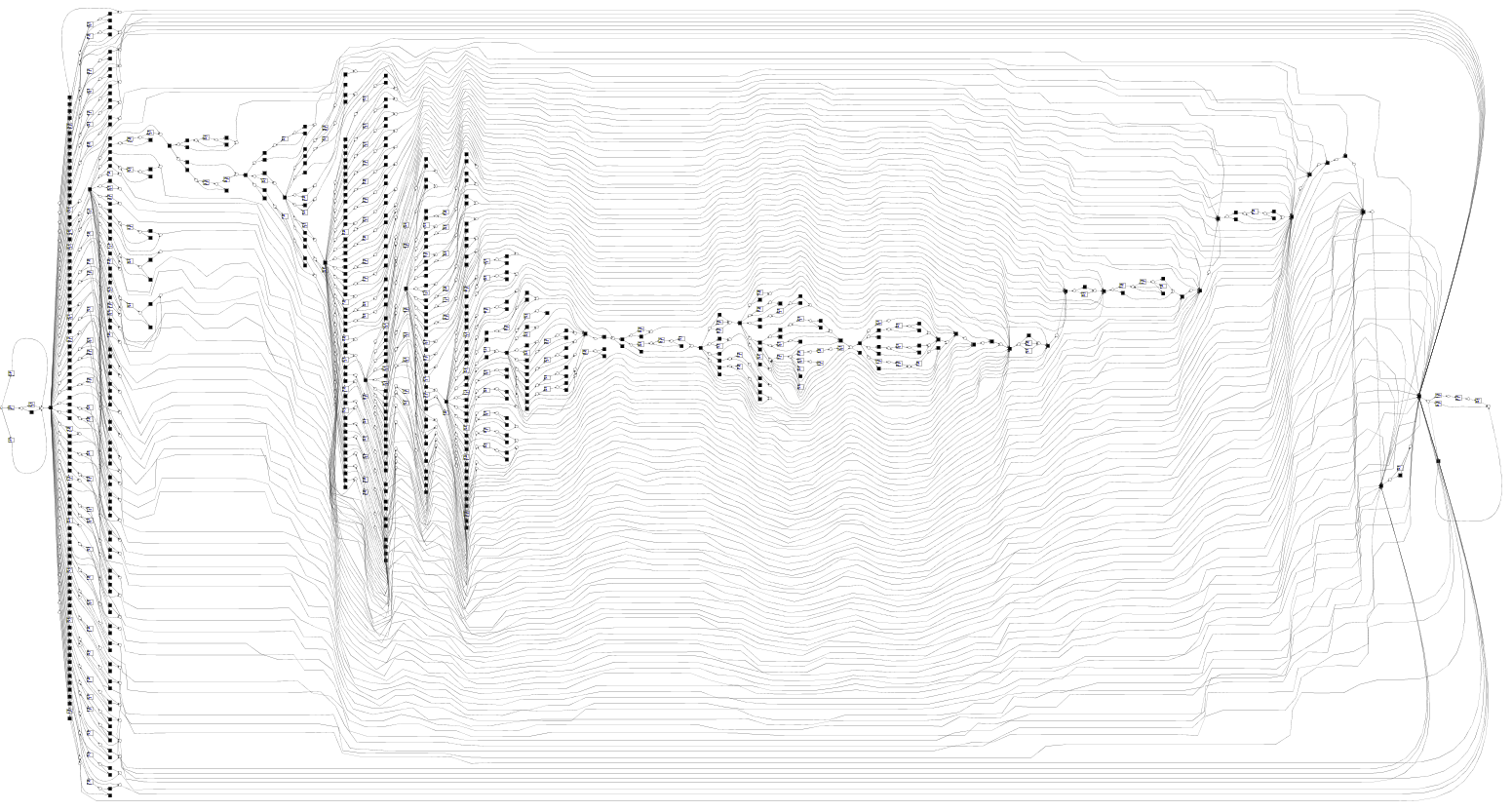}
	\caption{Red agent Petri-net generated by inductive miner algorithm (Fixed Simulation Depth, Fixed Minimax Search Depth, Iteration Times = 2000)}
	\label{fig: red_iteration_2000_inductive}
\end{figure}

In MCTS, each node stores a game state, and every available action serves as a branching factor. This architecture implies that a vast state space must be navigated during the expansion stage. Without effective pruning to eliminate redundant branching factors, a finite number of iterations may fail to fully expand a selected node within strict time constraints. If a node is not fully expanded before the computational budget is exhausted, the UCT algorithm cannot be properly applied to select the most optimal branching factor.To address this, we propose a pruning strategy utilizing a hashtable, an efficient data structure for rapid mapping. In a single MCTS iteration, once a node is selected for expansion, an auxiliary reward mechanism is employed to score each potential action. For example, if 12 actions ($A, B, C, D, E, F, G, H, I, J, K, L$) are retrieved, the mechanism assigns scores as follows: actions $A, B, C$ receive 10 points; $D, E$ receive 6 points; $F, G, H$ receive 4 points; and $I, J, K, L$ receive 0 points.We define the reward point as the key and the list of corresponding actions as the value within the hashtable. Figure~\ref{Pruning operation} and Table~\ref{Reward Hashtable} illustrate this proposed pruning methodology. By returning only the list of actions associated with the highest reward points, we effectively prune sub-optimal branches. A primary advantage of the hashtable is its support for near-instantaneous insertion, search, and deletion. Since the average time and space complexity for these operations is constant $O(1)$, implementing a hashtable for pruning does not introduce significant computational overhead to the MCTS model.

\begin{table}[!h]
	\centering
	\begin{tabular}{|c|c|}
		\hline
		\textbf{Reward} & \textbf{Actions} \\
		\hline
		10 & A, B, C \\
		\hline
		6 & D, E \\
		\hline
		4 & F, G, H \\
		\hline
		0 & I, J, K, L \\
		\hline
	\end{tabular}
	\caption{Reward hashtable}
	\label{Reward Hashtable}
\end{table}

\section{Process Models Figures}

\begin{figure}[!h]
	\centering
	\includegraphics[width=1\linewidth]{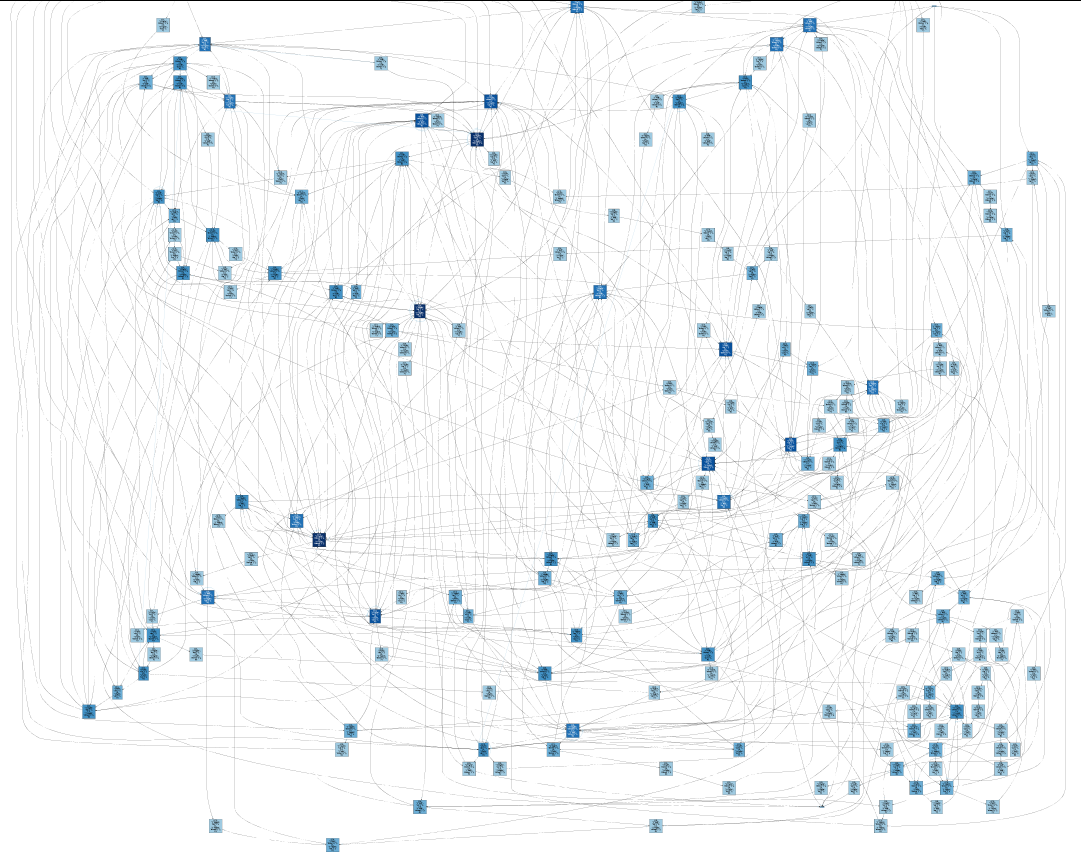}
	\caption{White agent C-net generated by iDHM (Fixed Simulation Depth, Fixed Minimax Search Depth, Iteration Times = 2000)}
	\label{fig: white_iteration_2000_iDHM}
\end{figure}

\begin{figure}[!h]
	\centering
	\includegraphics[width=1\linewidth]{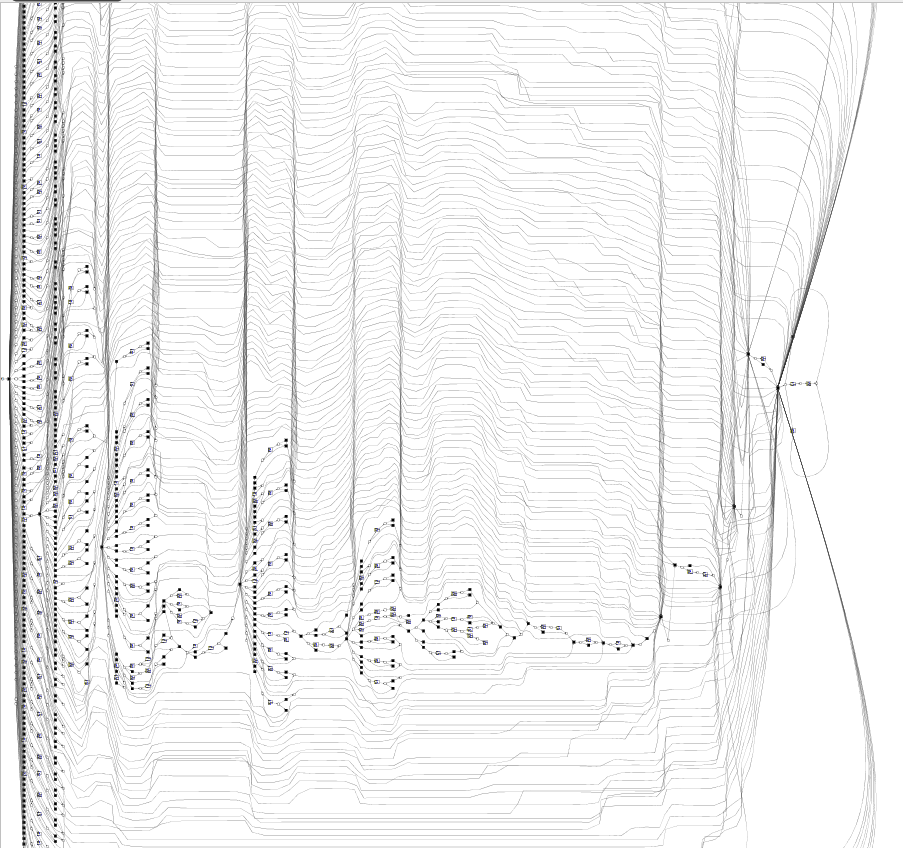}
	\caption{White agent Petri-net generated by inductive miner algorithm (Fixed Simulation Depth, Fixed Minimax Search Depth, Iteration Times = 2000)}
	\label{fig: white_iteration_2000_inductive}
\end{figure}

\begin{figure}[!h]
	\includegraphics[width=1\linewidth]{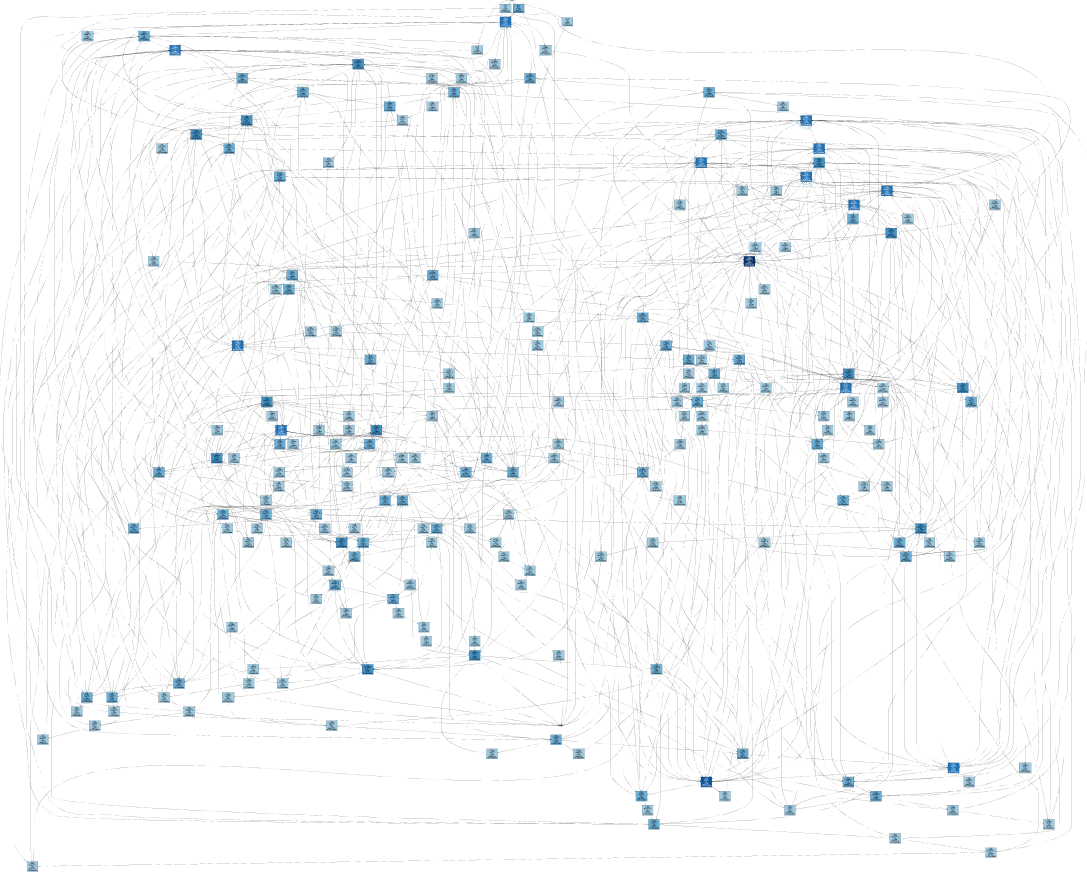}
	\caption{Red agent C-net generated by iDHM (Fixed Simulation Depth, Fixed Minimax Search Depth, Iteration Times = 3000)}
	\label{fig: red_iteration_3000_iDHM}
\end{figure}

\begin{figure}[!h]
	\centering
	\includegraphics[width=1\linewidth]{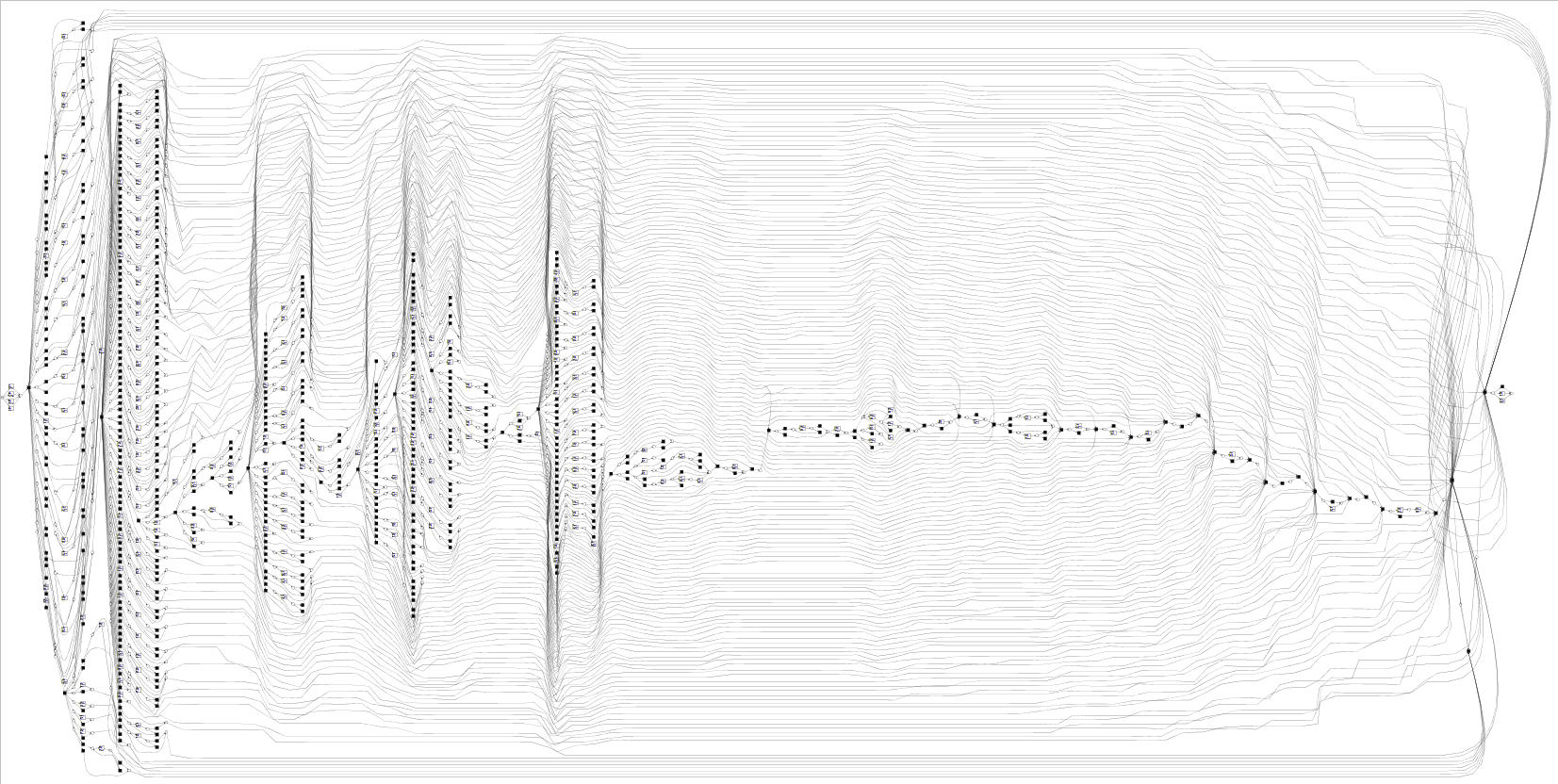}
	\caption{Red agent Petri-net generated by inductive miner algorithm (Fixed Simulation Depth, Fixed Minimax Search Depth, Iteration Times = 3000)}
	\label{fig: red_iteration_3000_inductive}
\end{figure}

\begin{figure}[!h]
	\centering
	\includegraphics[width=1\linewidth]{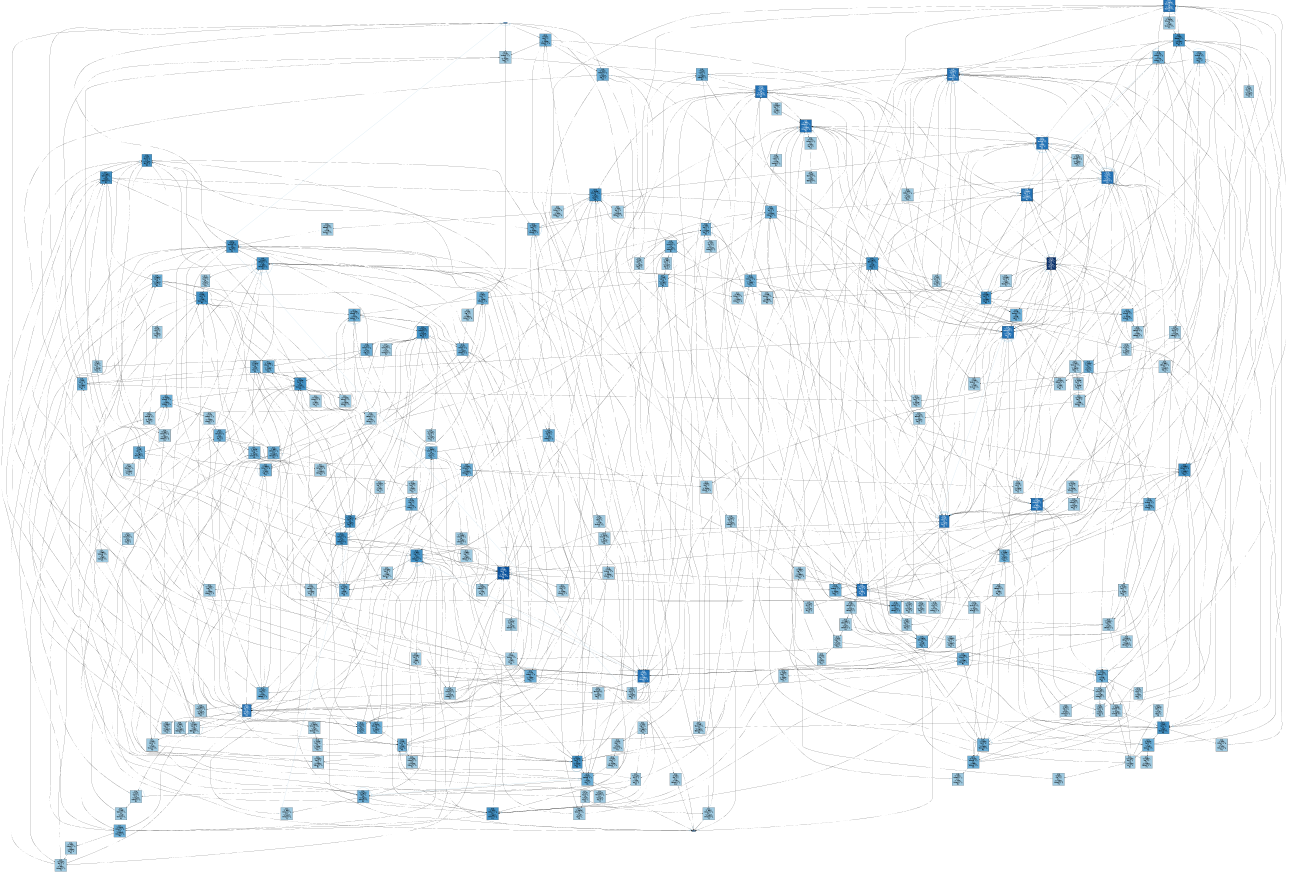}
	\caption{White Agent C-net generated by iDHM (Fixed Simulation Depth, Fixed Minimax Search Depth, Iteration Times = 3000)}
	\label{fig: white_iteration_3000_iDHM}
\end{figure}

\begin{figure}[!h]
	\centering
	\includegraphics[width=1\linewidth]{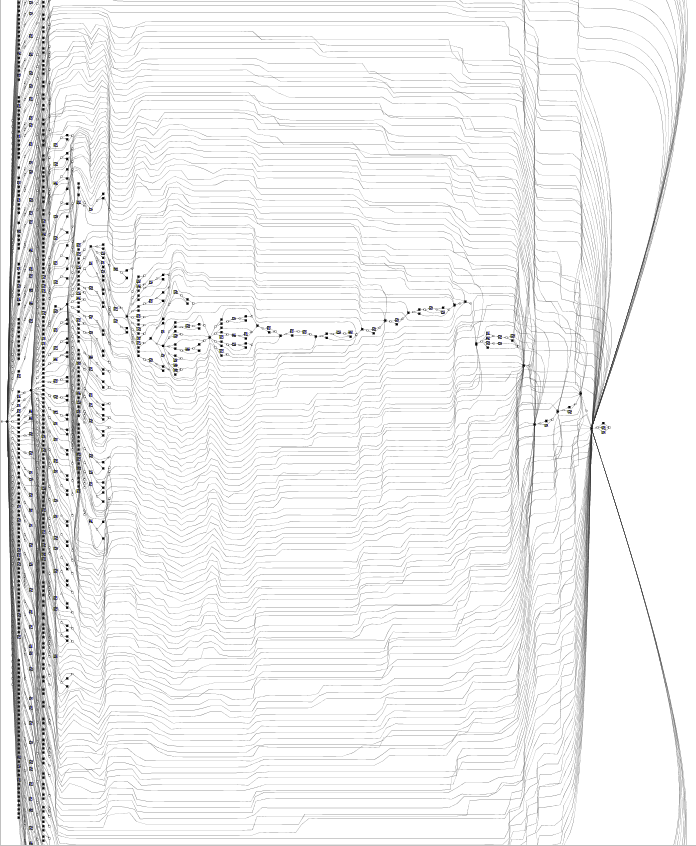}
	\caption{White Agent Petri-net generated by inductive miner algorithm (Fixed Simulation Depth, Fixed Minimax Search Depth, Iteration Times = 3000)}
	\label{fig: white_iteration_3000_inductive}
\end{figure}

\begin{figure}[!h]
	\centering
	\includegraphics[width=1\linewidth]{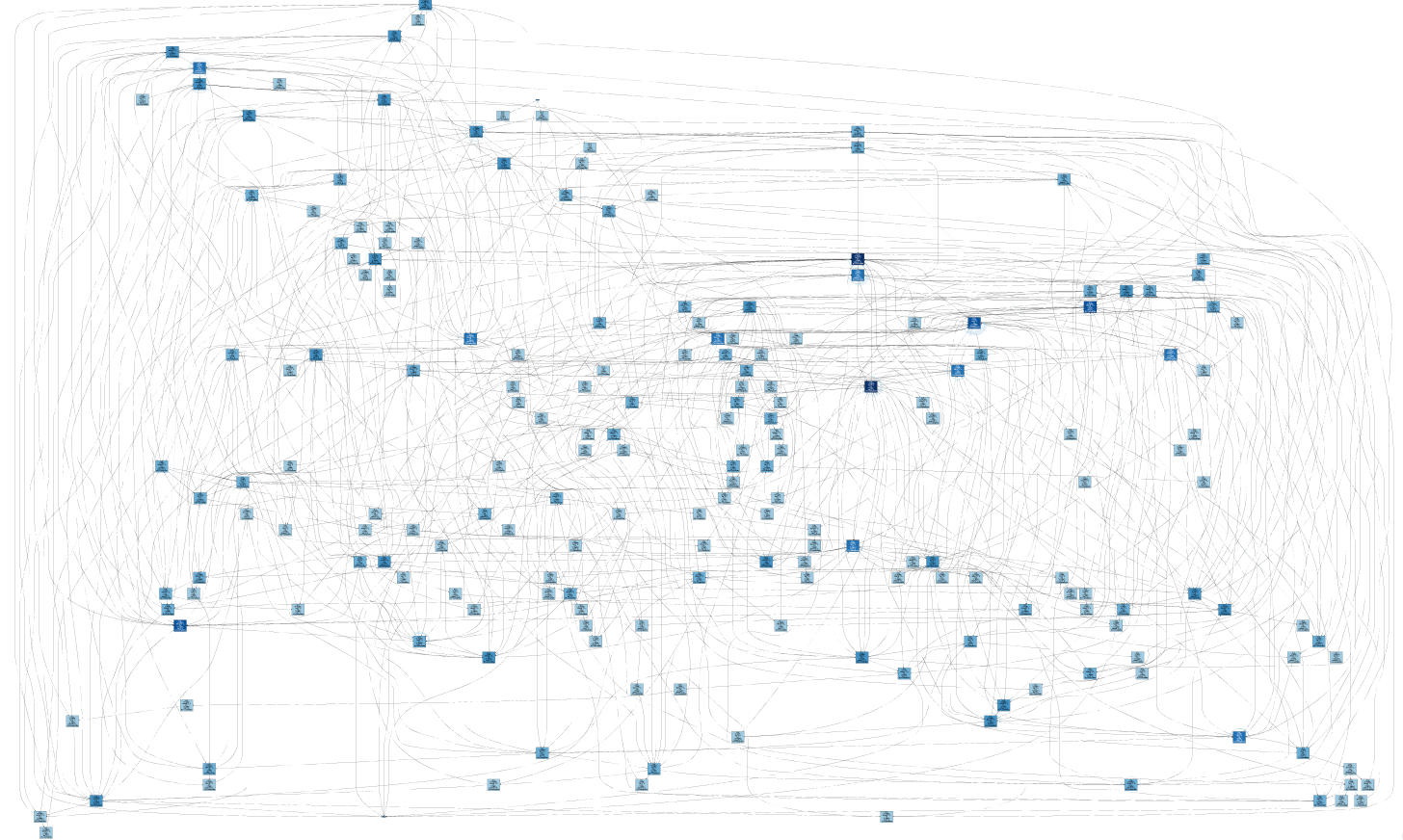}
	\caption{Red Agent C-net generated by iDHM (Fixed Iteration Times, Fixed Minimax Search Depth, Simulation Depth = 10)}
	\label{fig: red_simulation_10_iDHM}
\end{figure}

\begin{figure}[!h]
	\centering
	\includegraphics[width=1\linewidth]{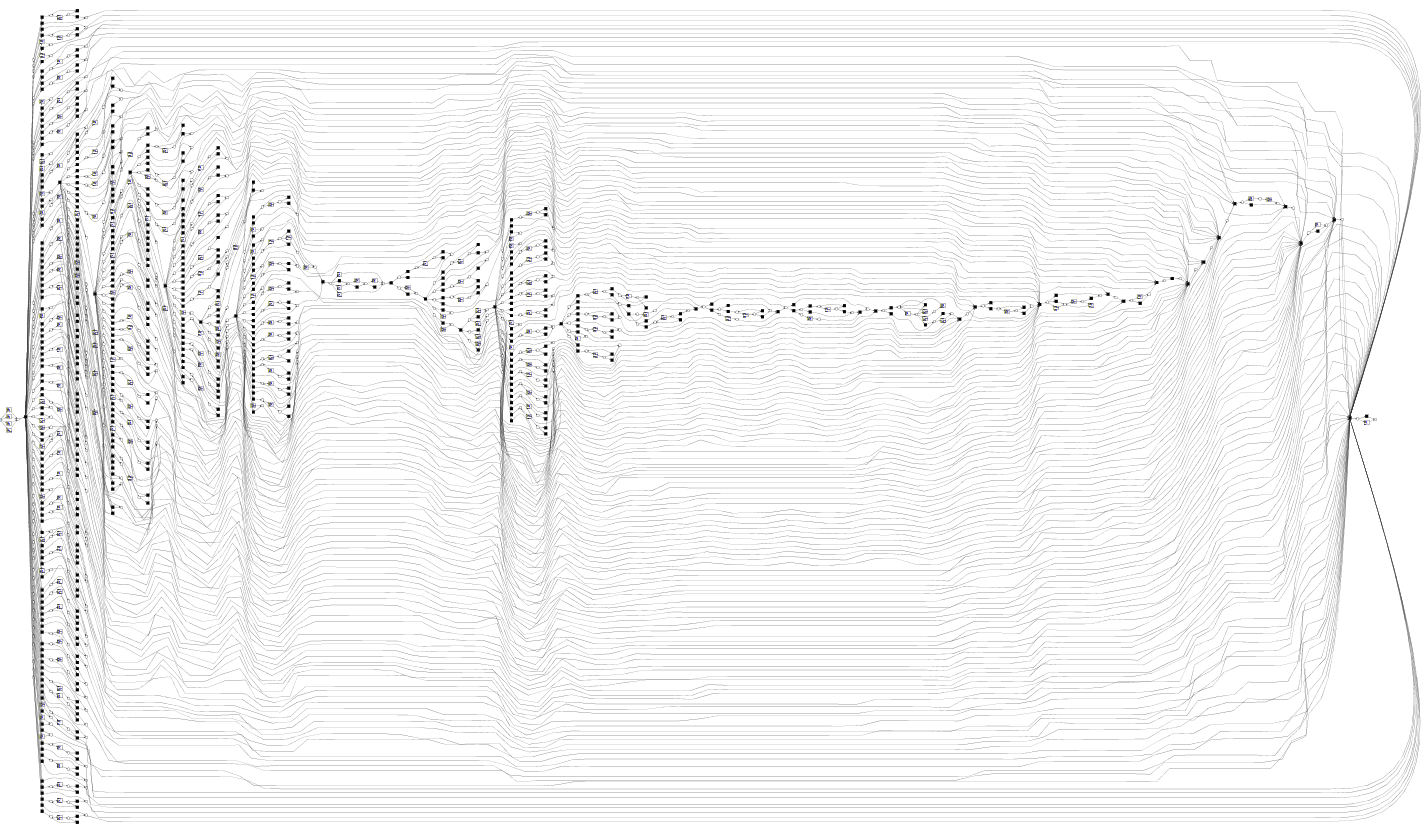}
	\caption{Red Agent Petri-net generated by inductive miner algorithm (Fixed Iteration Times, Fixed Minimax Search Depth, Simulation Depth = 10)}
	\label{fig: red_simulation_10_inductive}
\end{figure}

\begin{figure}[!h]
	\centering
	\includegraphics[width=1\linewidth]{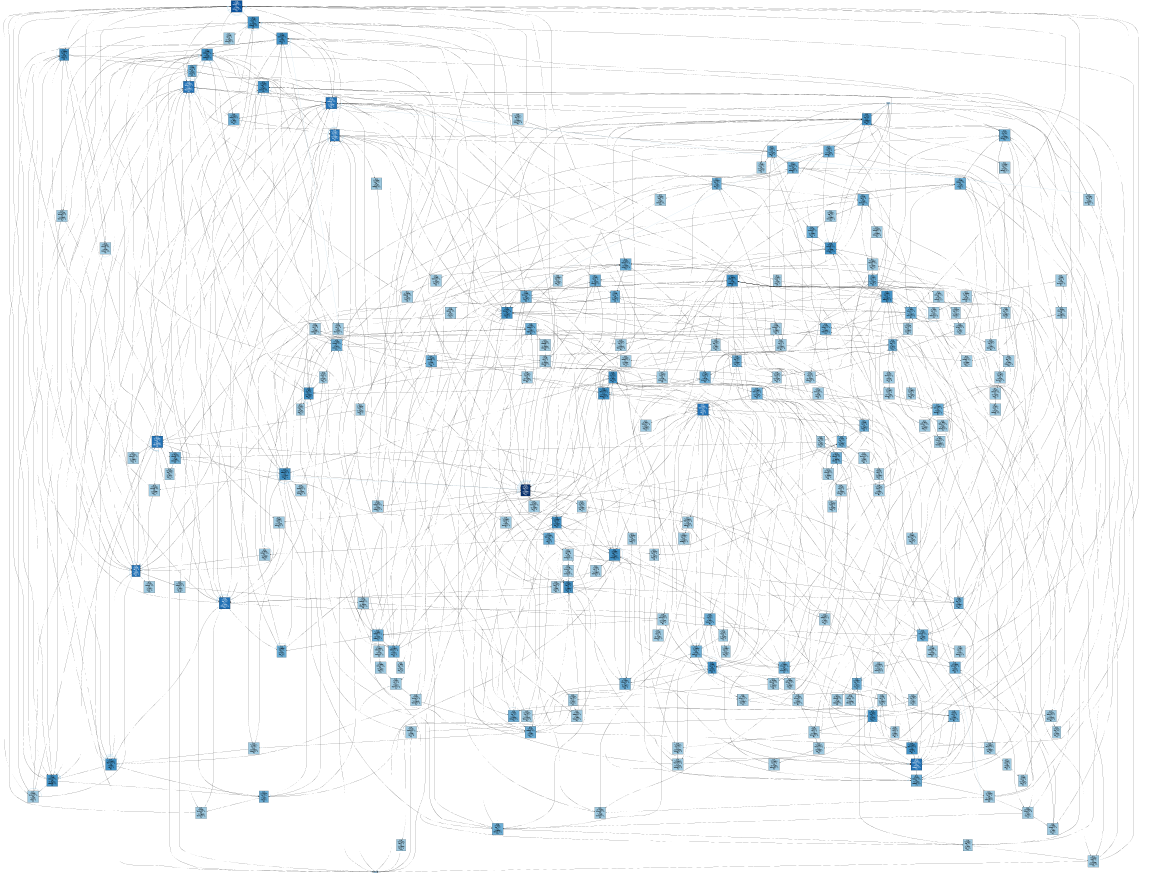}
	\caption{White Agent C-net generated by iDHM (Fixed Iteration Times, Fixed Minimax Search Depth, Simulation Depth = 10)}
	\label{fig: white_simulation_10_iDHM}
\end{figure}

\begin{figure}[!h]
	\centering
	\includegraphics[width=1\linewidth]{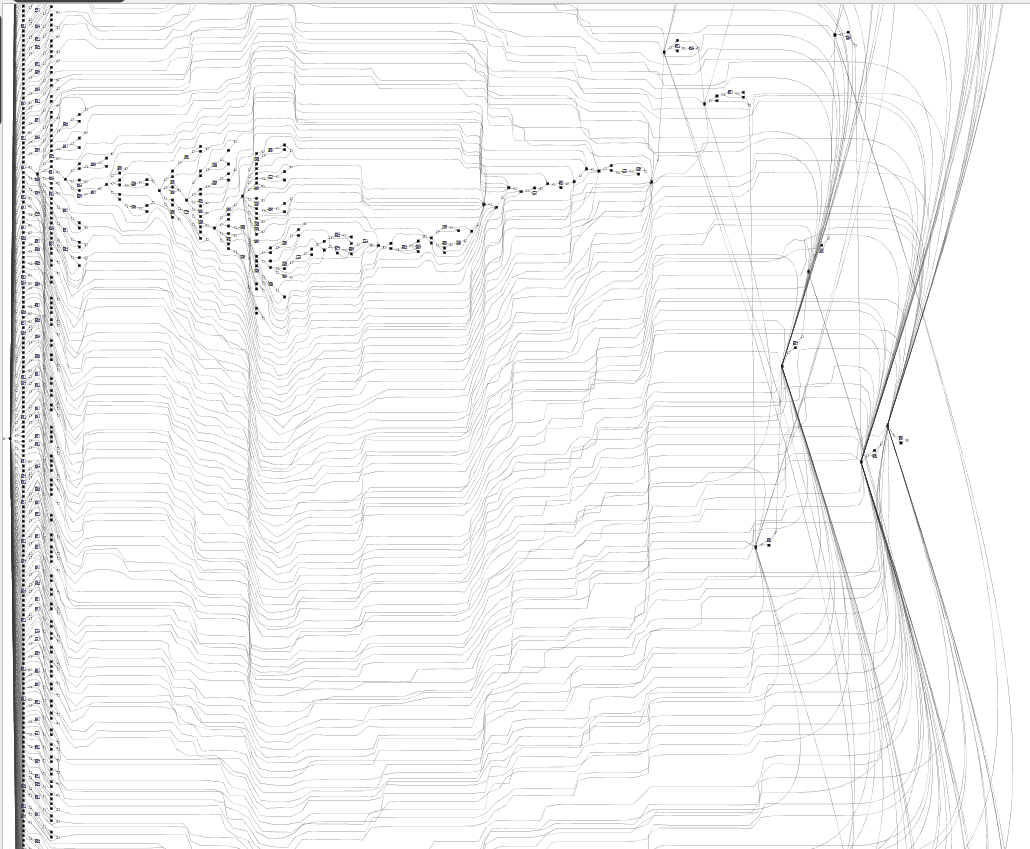}
	\caption{White Agent Petri-net generated by inductive miner algorithm (Fixed Iteration Times, Fixed Minimax Search Depth, Simulation Depth = 10)}
	\label{fig: white_simulation_10_inductive}
\end{figure}

\begin{figure}[!h]
	\centering
	\includegraphics[width=1\linewidth]{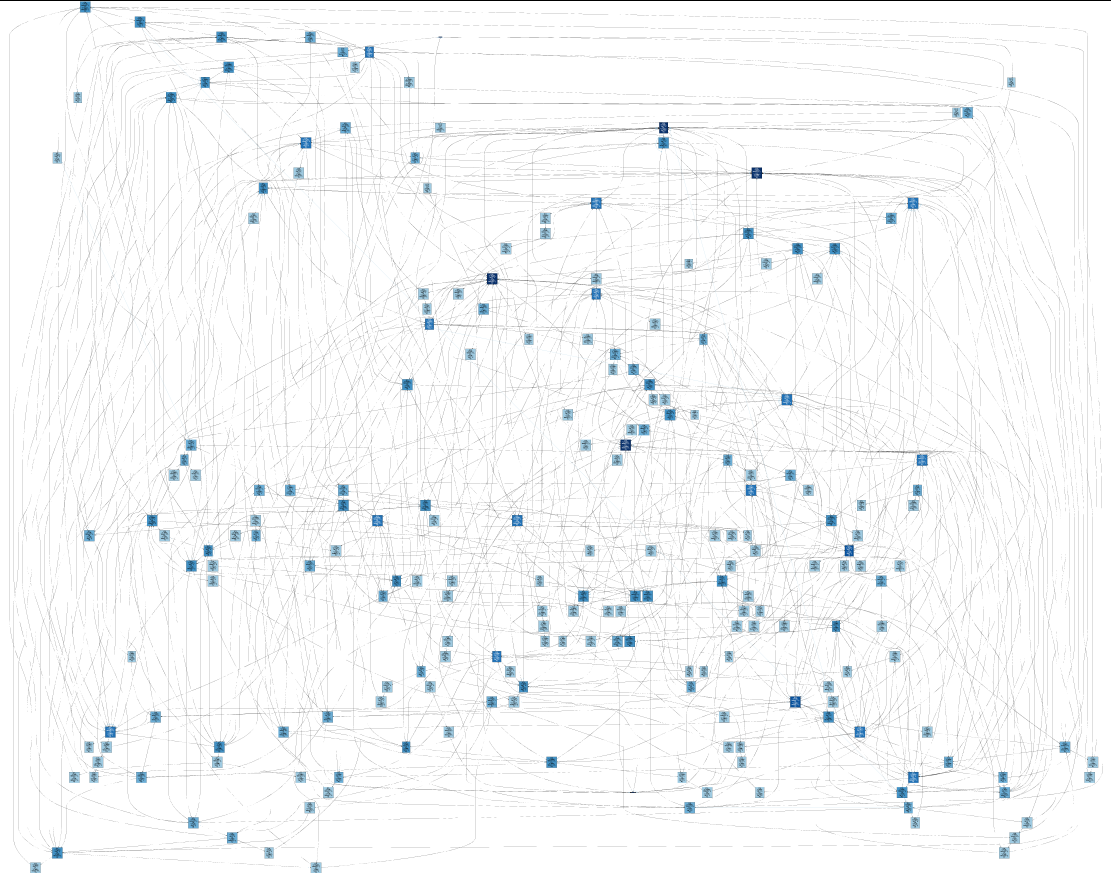}
	\caption{Red Agent C-net generated by iDHM (Fixed Iteration Times, Fixed Minimax Search Depth, Simulation Depth = 20)}
	\label{fig: red_simulation_20_iDHM}
\end{figure}

\begin{figure}[!h]
	\centering
	\includegraphics[width=1\linewidth]{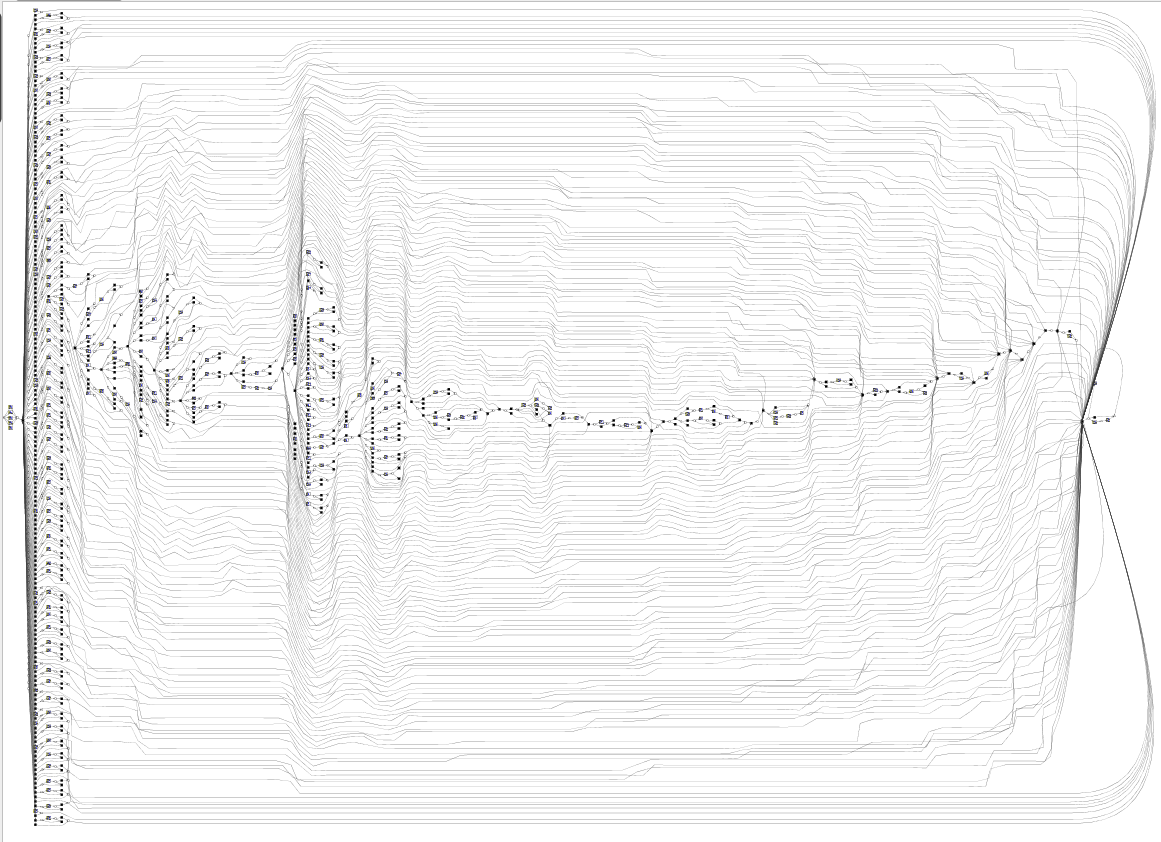}
	\caption{Red Agent Petri-net generated by inductive miner algorithm (Fixed Iteration Times, Fixed Minimax Search Depth, Simulation Depth = 20)}
	\label{fig: red_simulation_20_inductive}
\end{figure}

\begin{figure}[!h]
	\centering
	\includegraphics[width=1\linewidth]{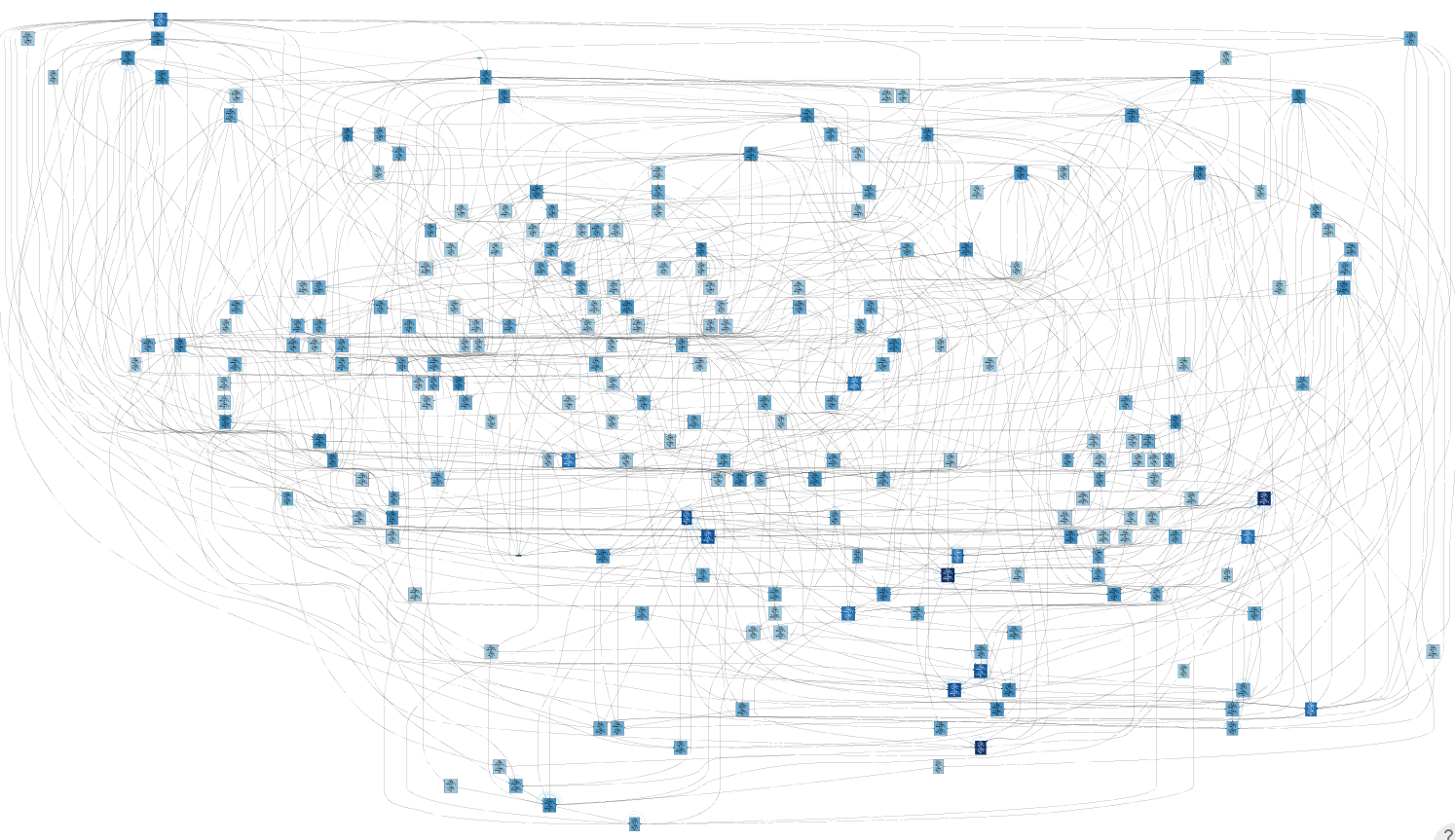}
	\caption{White Agent C-net generated by iDHM (Fixed Iteration Times, Fixed Minimax Search Depth, Simulation Depth = 20)}
	\label{fig: white_simulation_20_iDHM}
\end{figure}

\begin{figure}[!h]
	\centering
	\includegraphics[width=1\linewidth]{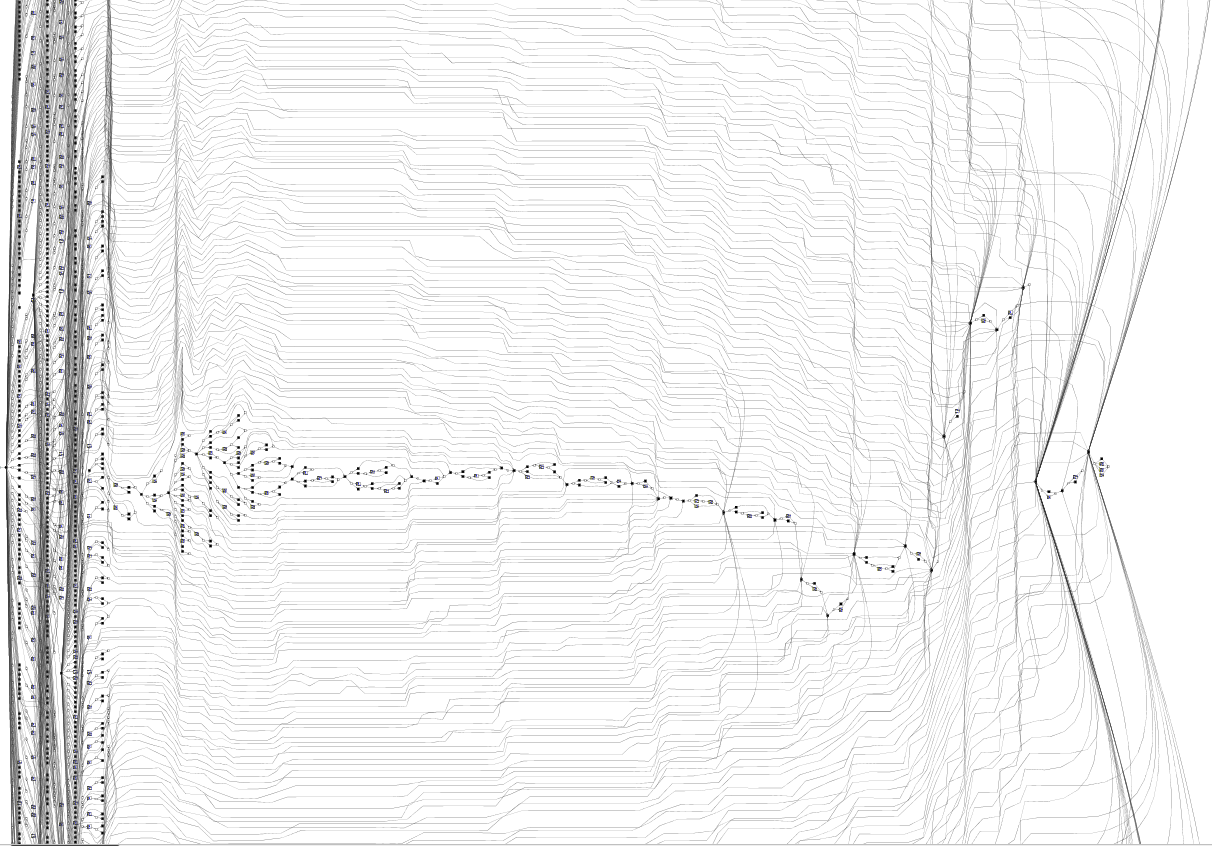}
	\caption{White Agent Petri-net generated by inductive miner algorithm (Fixed Iteration Times, Fixed Minimax Search Depth, Simulation Depth = 20)}
	\label{fig: white_simulation_20_inductive}
\end{figure}

\begin{figure}[!h]
	\centering
	\includegraphics[width=1\linewidth]{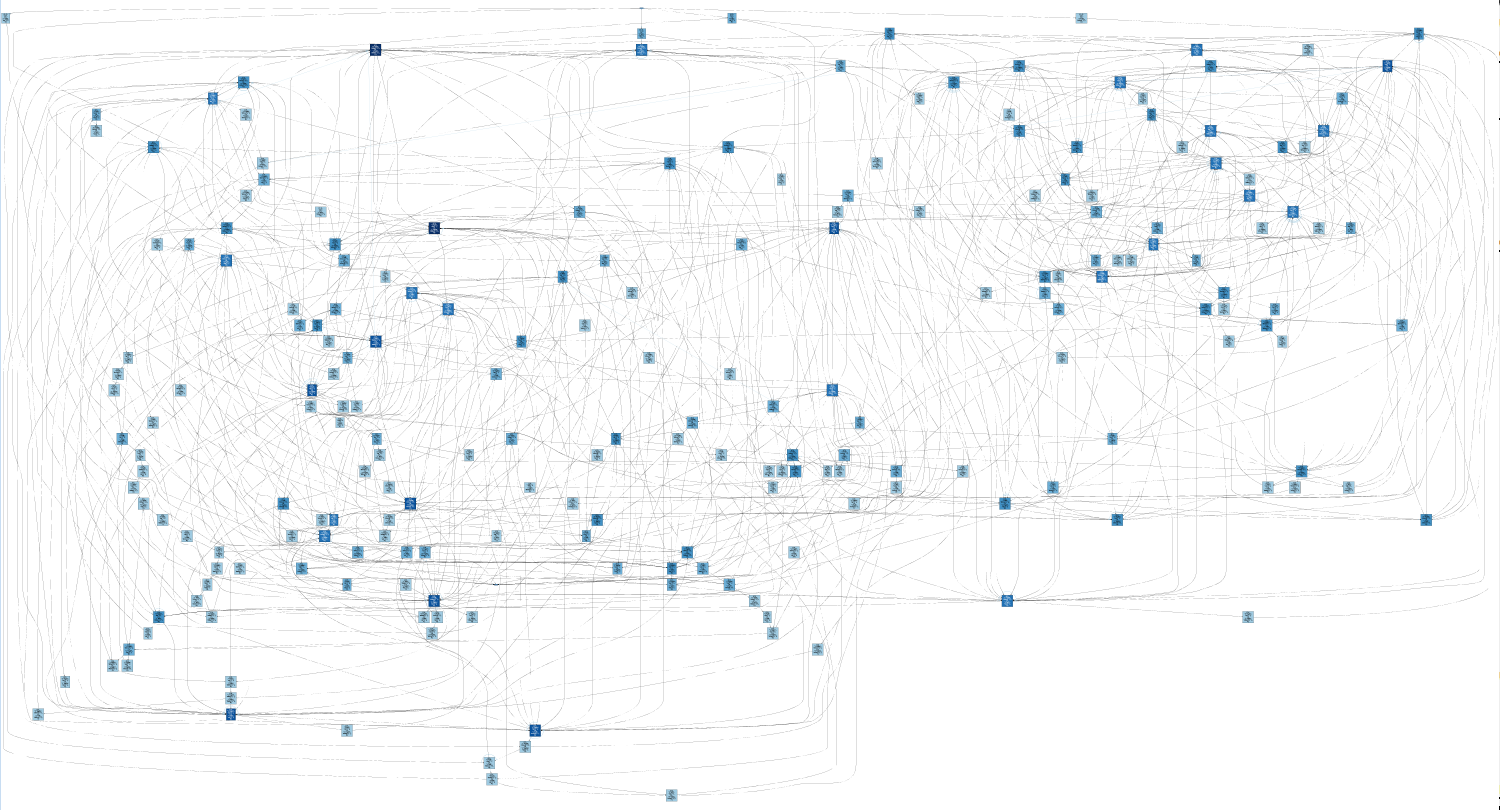}
	\caption{Red agent C-net generated by iDHM (Fixed Iteration Times, Fixed Minimax Search Depth, Simulation Depth = 30)}
	\label{fig: red_simulation_30_iDHM}
\end{figure}

\begin{figure}[!h]
	\centering
	\includegraphics[width=1\linewidth]{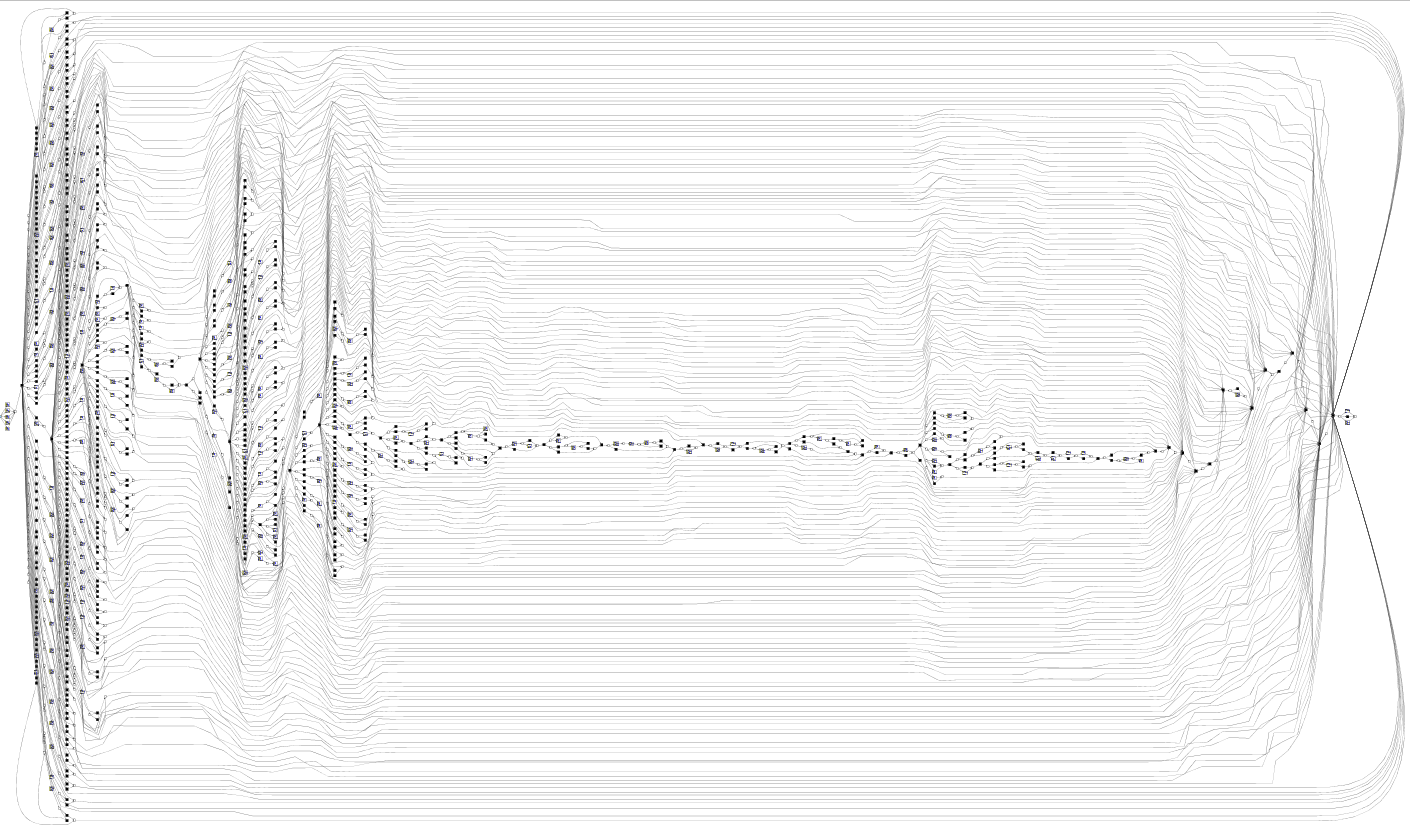}
	\caption{Red agent Petri-net generated by inductive miner algorithm (Fixed Iteration Times, Fixed Minimax Search Depth, Simulation Depth = 30)}
	\label{fig: red_simulation_30_inductive}
\end{figure}

\begin{figure}[!h]
	\centering
	\includegraphics[width=1\linewidth]{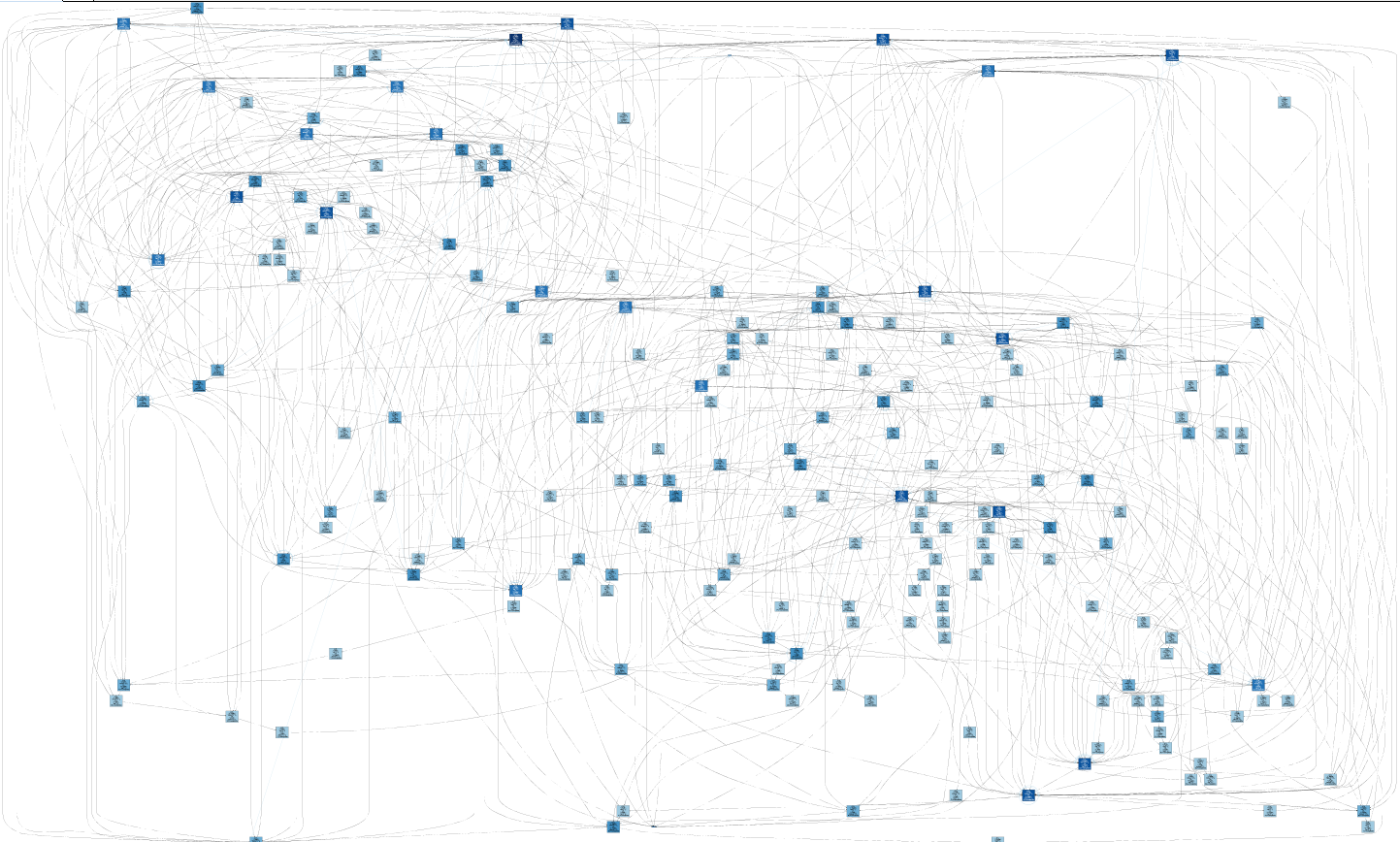}
	\caption{White agent C-net generated by iDHM (Fixed Iteration Times, Fixed Minimax Search Depth, Simulation Depth = 30)}
	\label{fig: white_simulation_30_iDHM}
\end{figure}

\begin{figure}[!h]
	\centering
	\includegraphics[width=1\linewidth]{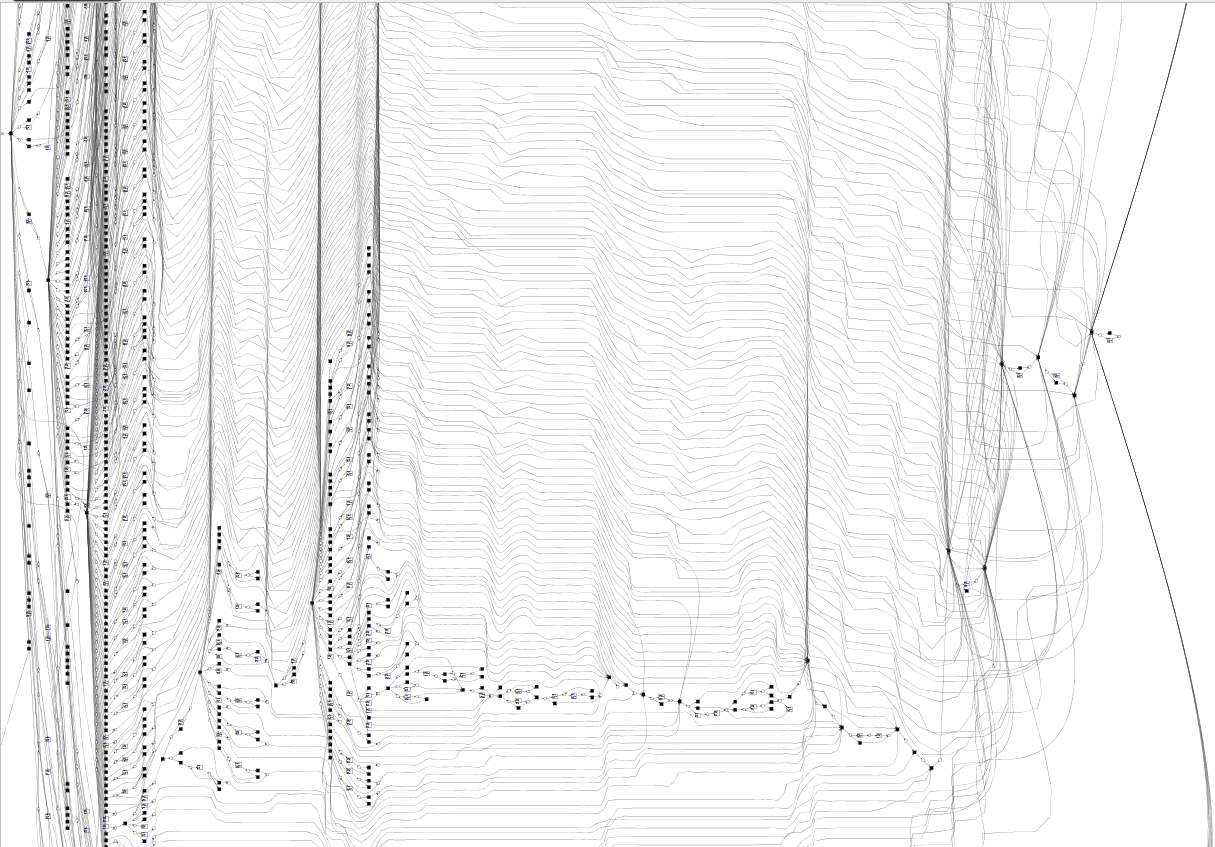}
	\caption{White agent Petri-net generated by inductive miner algorithm (Fixed Iteration Times, Fixed Minimax Search Depth, Simulation Depth = 30)}
	\label{fig: white_simulation_30_inductive}
\end{figure}

\begin{figure}[!h]
	\centering
	\includegraphics[width=1\linewidth]{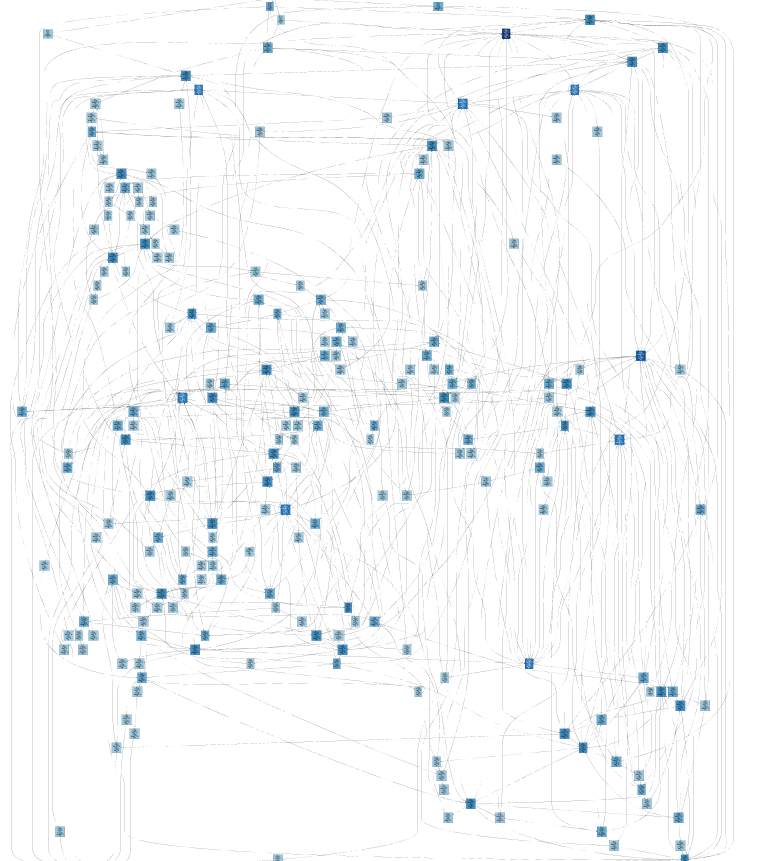}
	\caption{Red agent C-net generated by iDHM (Fixed Simulation Depth, Fixed Iteration Times, Minimax Search Depth = 1)}
	\label{fig: red_minimax_1_iDHM}
\end{figure}

\begin{figure}[!h]
	\centering
	\includegraphics[width=1\linewidth]{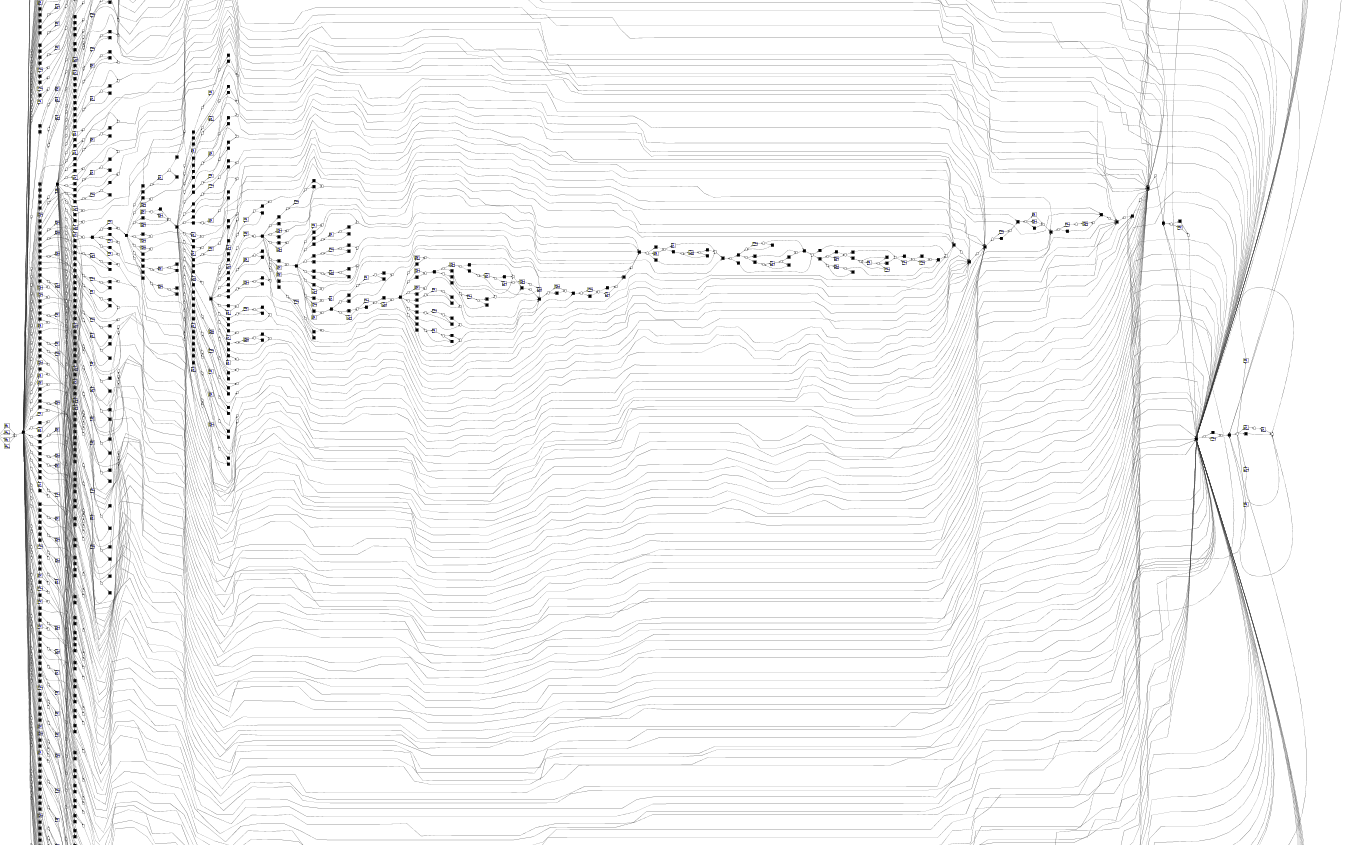}
	\caption{Red agent Petri-net generated by inductive miner algorithm (Fixed Simulation Depth, Fixed Iteration Times, Minimax Search Depth = 1)}
	\label{fig: red_minimax_1_inductive}
\end{figure}

\begin{figure}[!h]
	\centering
	\includegraphics[width=1\linewidth]{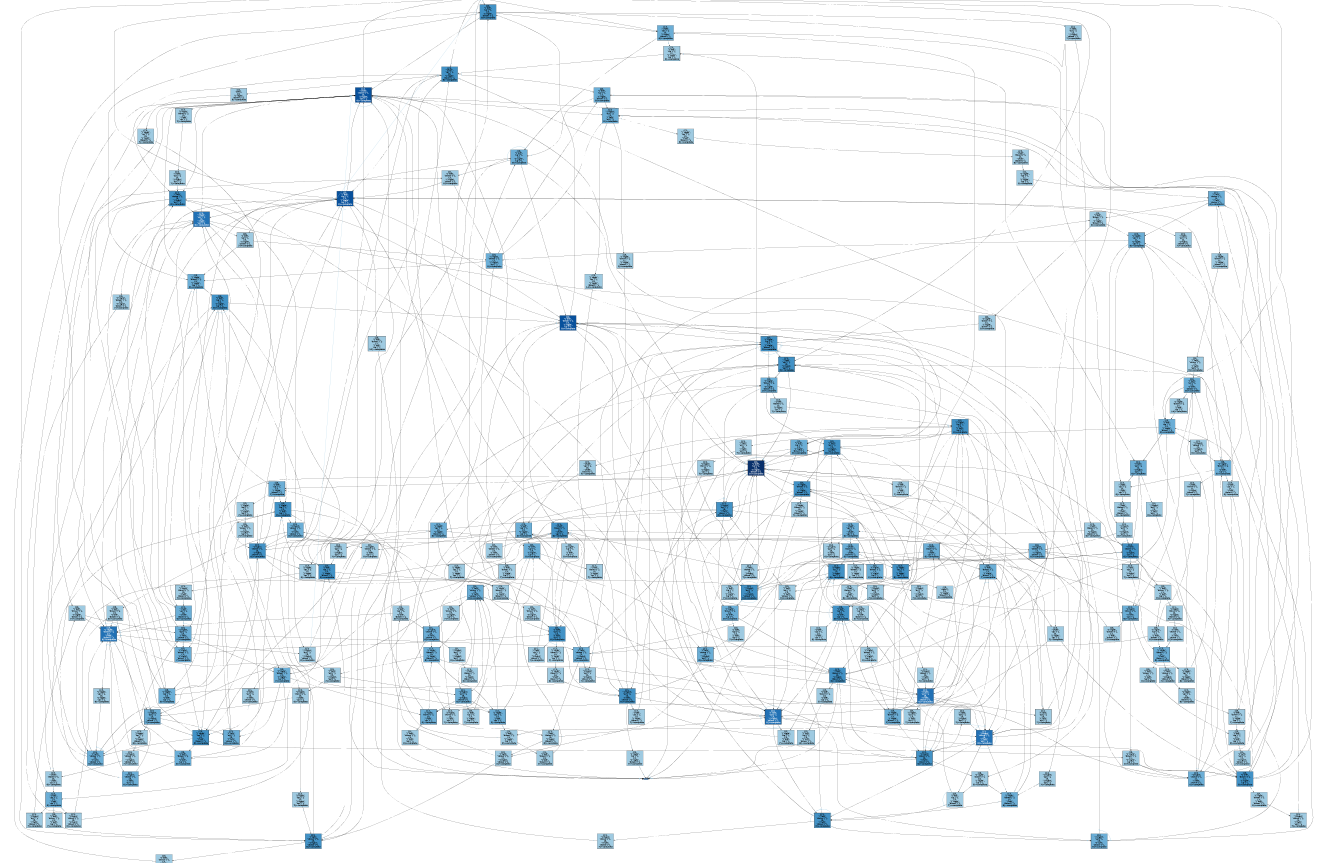}
	\caption{White agent C-net generated by iDHM (Fixed Simulation Depth, Fixed Iteration Times, Minimax Search Depth = 1)}
	\label{fig: white_minimax_1_iDHM}
\end{figure}

\begin{figure}[!h]
	\centering
	\includegraphics[width=1\linewidth]{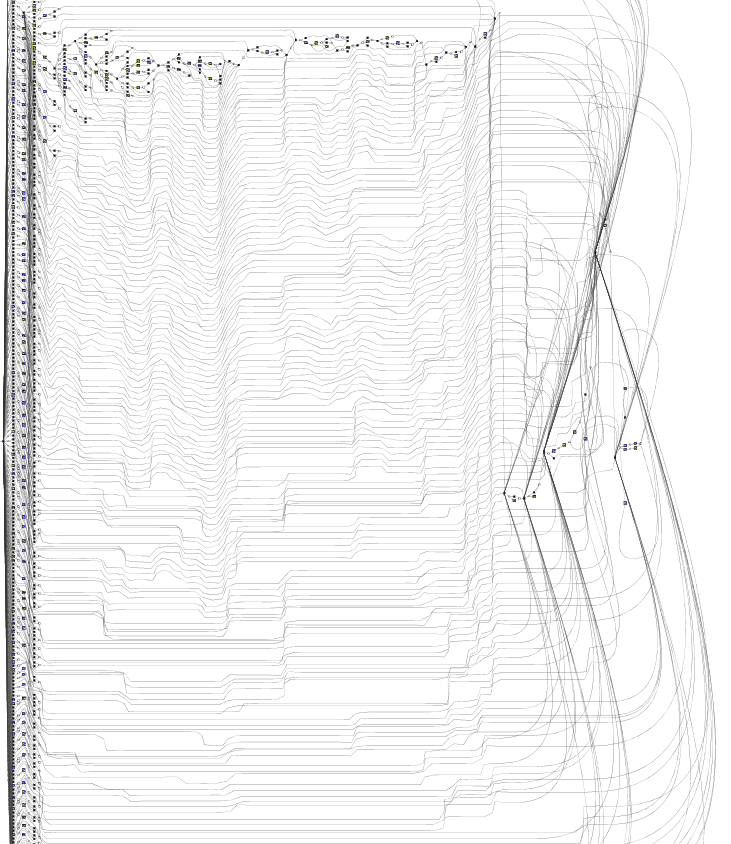}
	\caption{White agent Petri-net generated by inductive miner algorithm (Fixed Simulation Depth, Fixed Iteration Times, Minimax Search Depth = 1)}
	\label{fig: white_minimax_1_inductive}
\end{figure}

\begin{figure}[!h]
	\centering
	\includegraphics[width=1\linewidth]{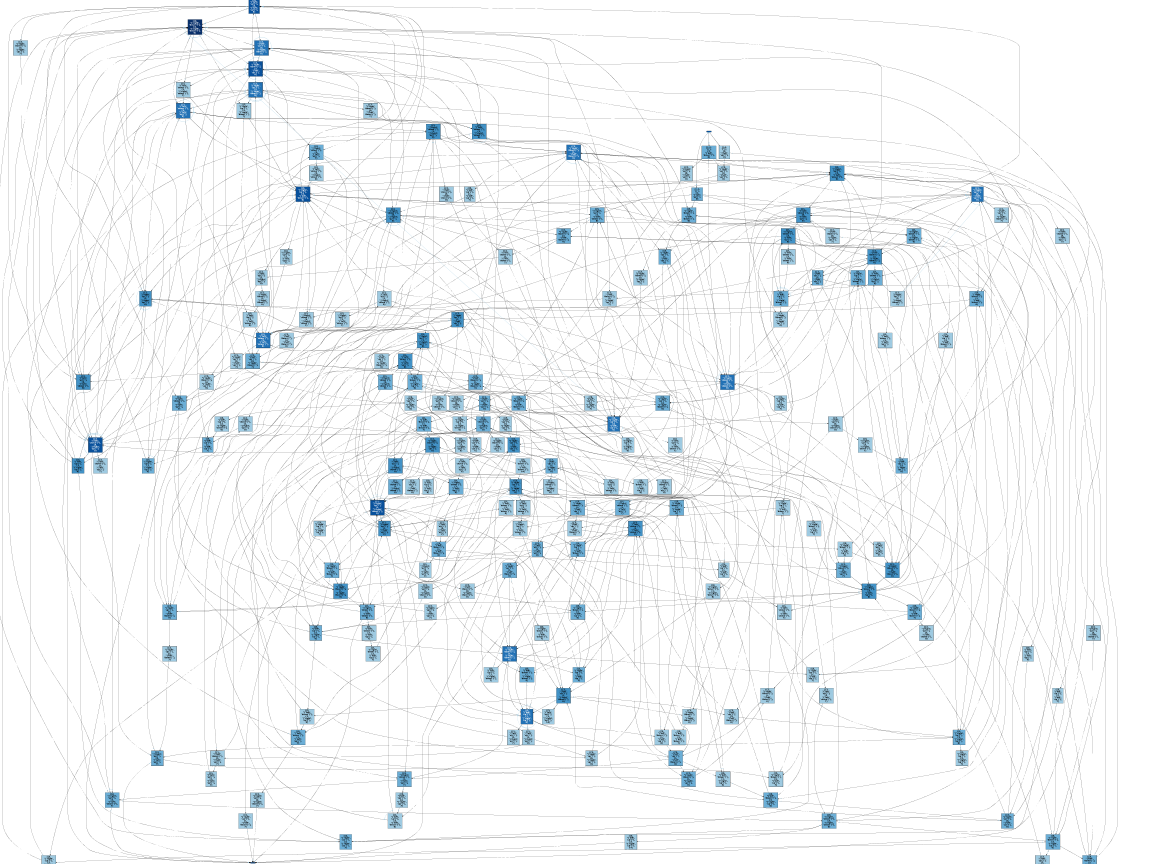}
	\caption{Red agent C-net generated by iDHM (Fixed Simulation Depth, Fixed Iteration Times, Minimax Search Depth = 2)}
	\label{fig: red_minimax_2_iDHM}
\end{figure}

\begin{figure}[!h]
	\centering
	\includegraphics[width=1\linewidth]{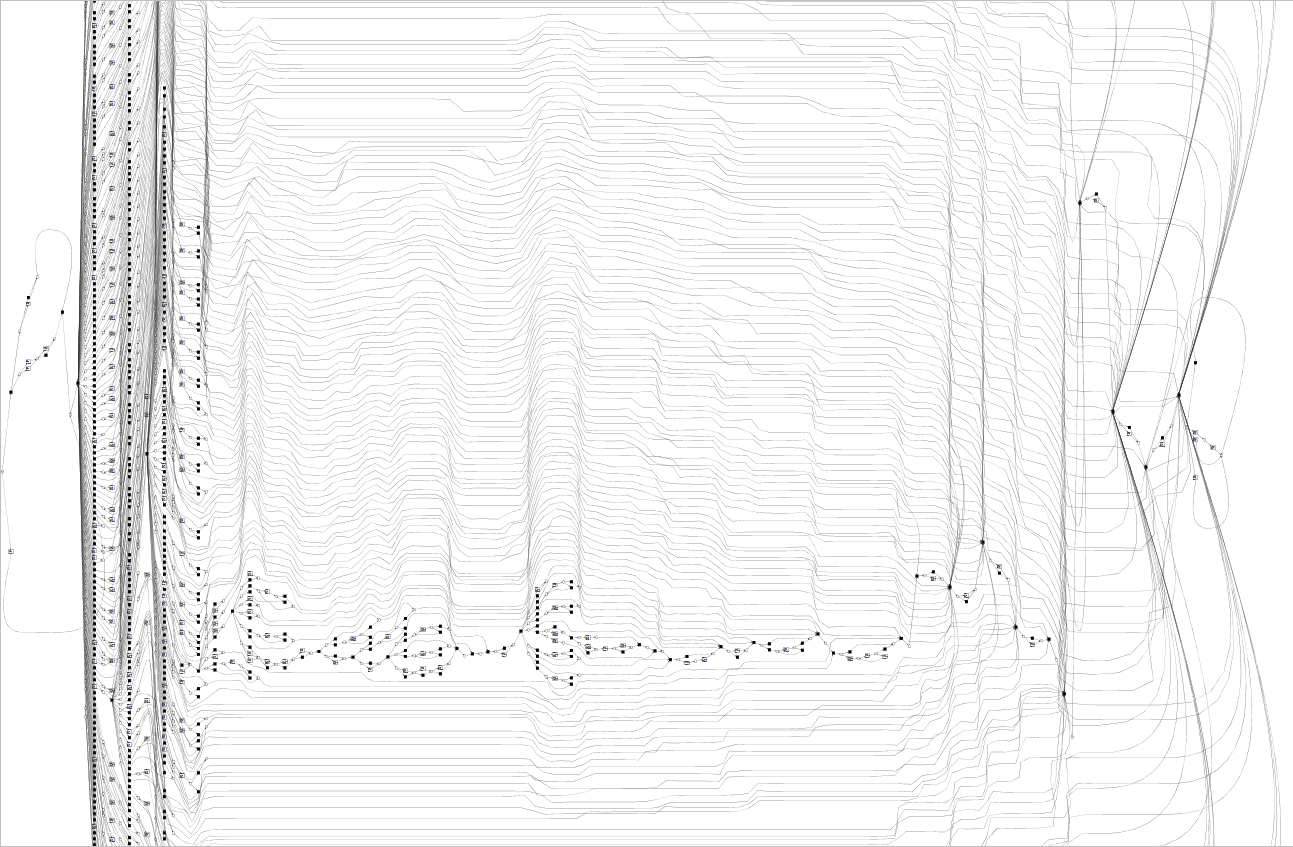}
	\caption{Red agent Petri-net generated by inductive miner algorithm (Fixed Simulation Depth, Fixed Iteration Times, Minimax Search Depth = 2)}
	\label{fig: red_minimax_2_inductive}
\end{figure}

\begin{figure}[!h]
	\centering
	\includegraphics[width=1\linewidth]{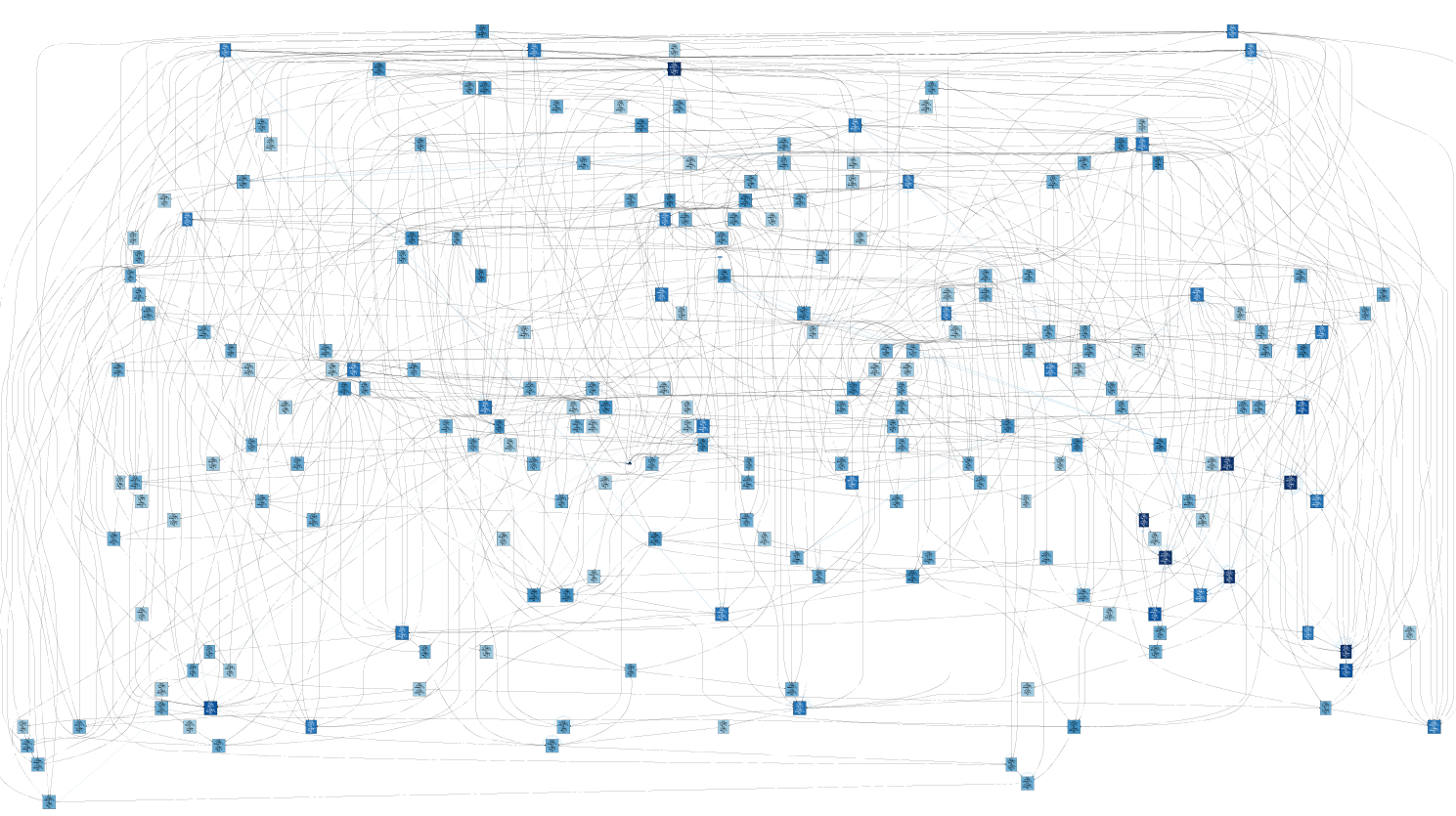}
	\caption{White agent C-net generated by iDHM (Fixed Simulation Depth, Fixed Iteration Times, Minimax Search Depth = 2)}
	\label{fig: white_minimax_2_iDHM}
\end{figure}

\begin{figure}[!h]
	\centering
	\includegraphics[width=1\linewidth]{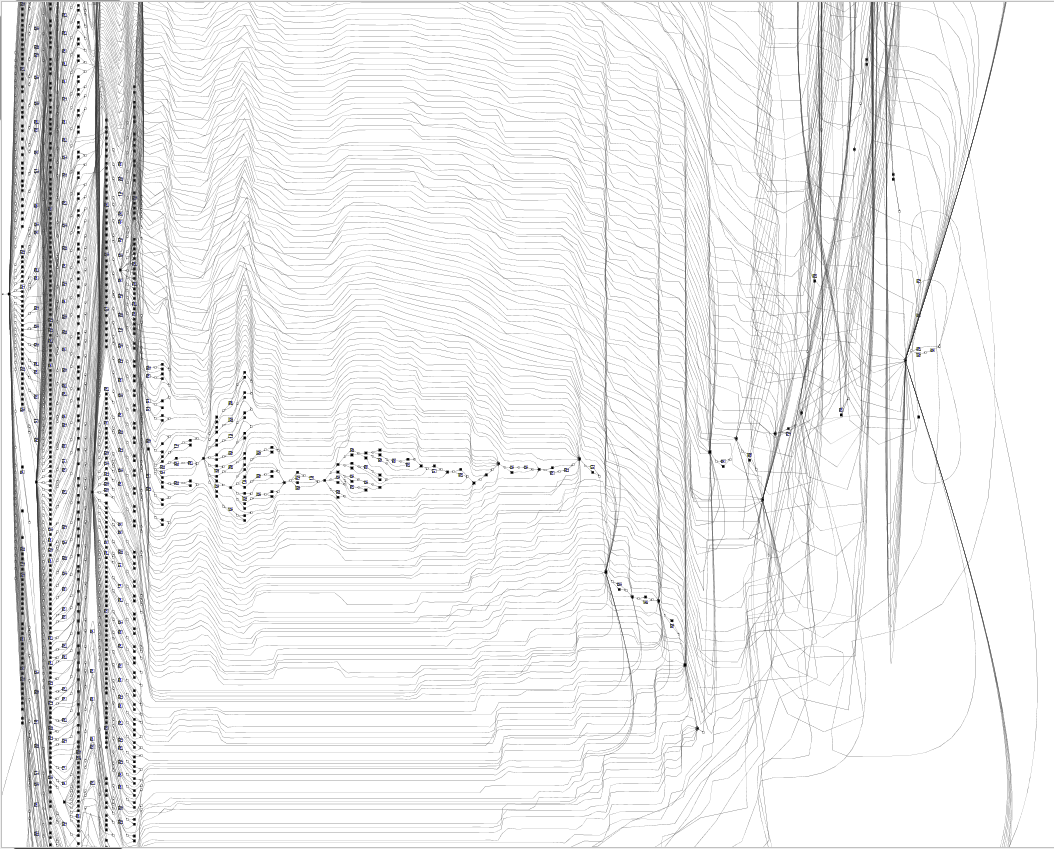}
	\caption{White agent Petri-net generated by inductive miner algorithm (Fixed Simulation Depth, Fixed Iteration Times, Minimax Search Depth = 2)}
	\label{fig: white_minimax_2_inductive}
\end{figure}

\begin{figure}[!h]
	\centering
	\includegraphics[width=1\linewidth]{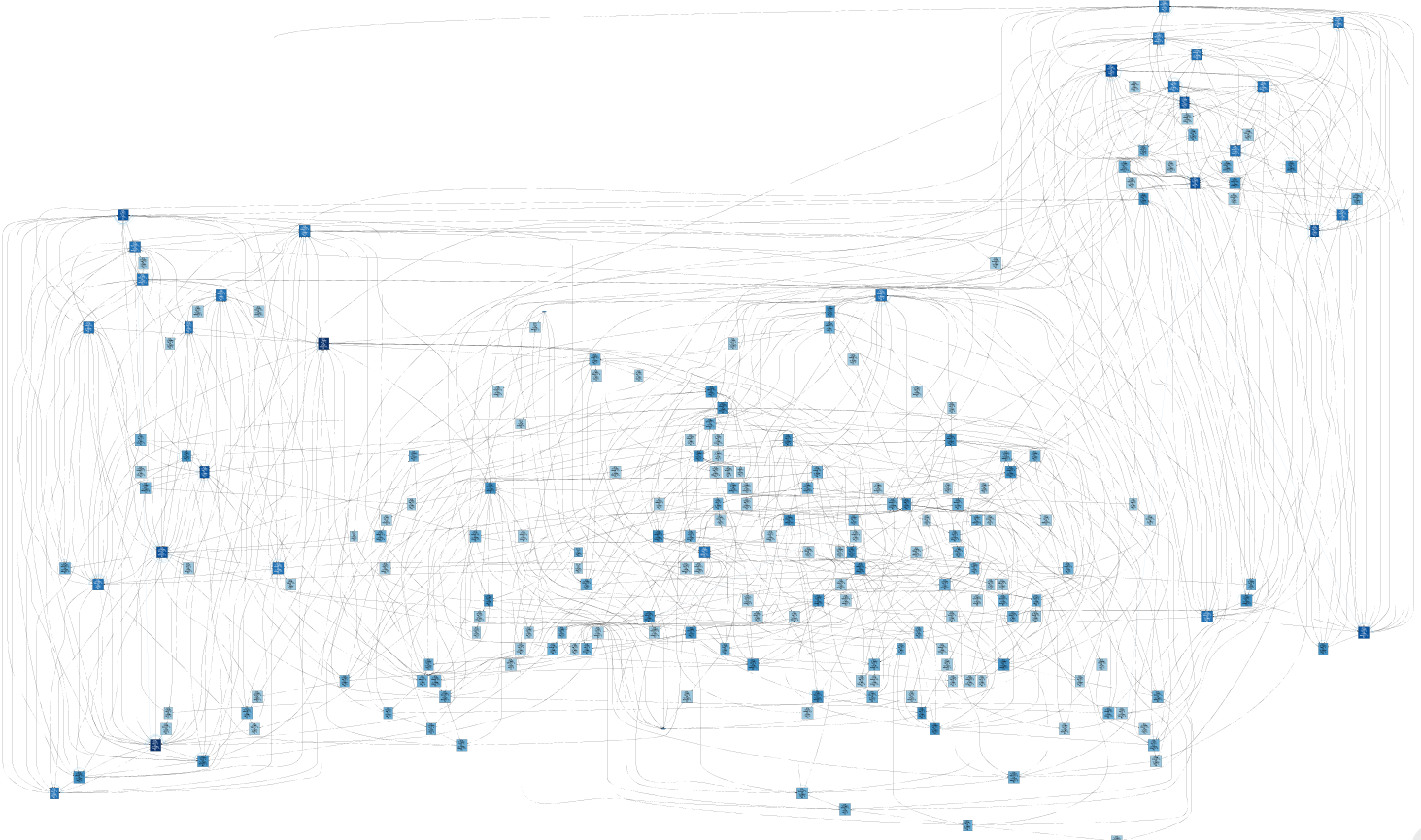}
	\caption{Red agent C-net generated by iDHM (Fixed Simulation Depth, Fixed Iteration Times, Minimax Search Depth = 3)}
	\label{fig: red_minimax_3_iDHM}
\end{figure}

\begin{figure}[!h]
	\centering
	\includegraphics[width=1\linewidth]{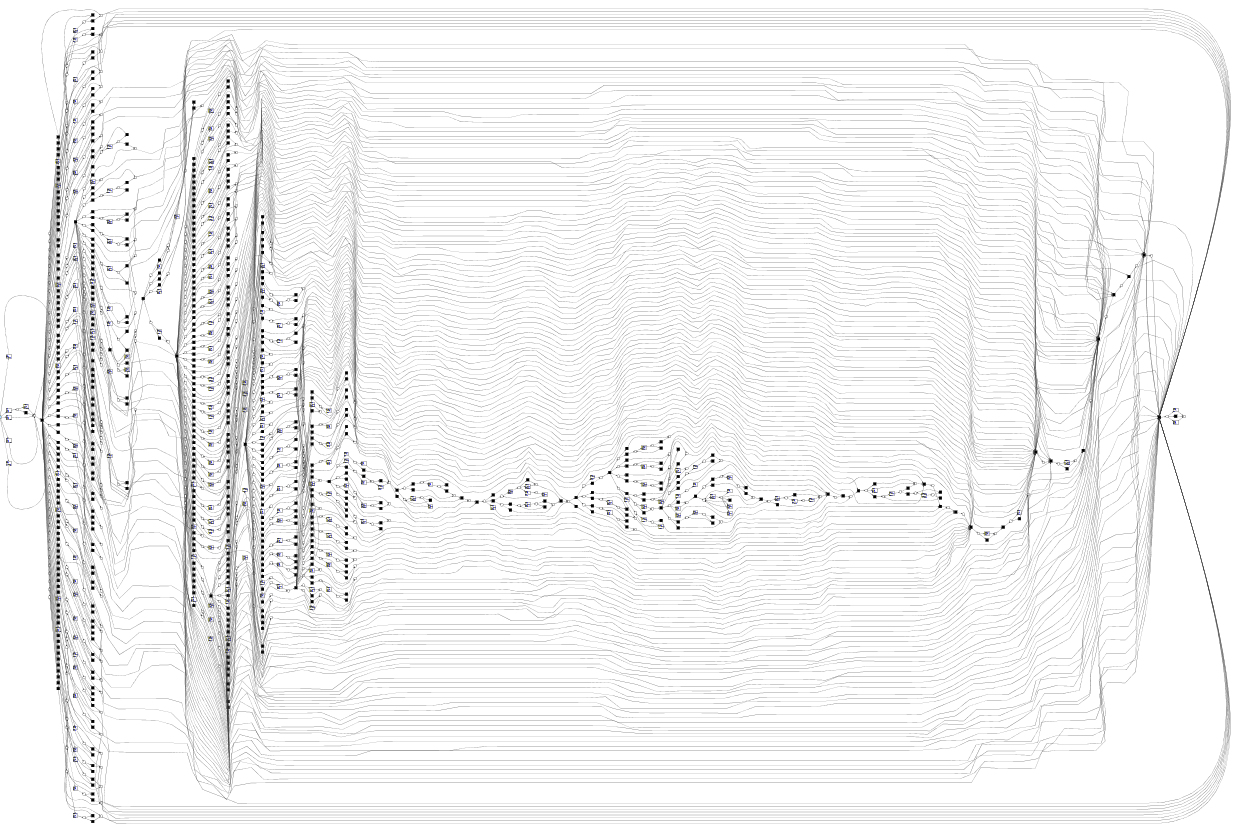}
	\caption{Red agent Petri-net generated by inductive miner algorithm (Fixed Simulation Depth, Fixed Iteration Times, Minimax Search Depth = 3)}
	\label{fig: red_minimax_3_inductive}
\end{figure}

\begin{figure}[!h]
	\centering
	\includegraphics[width=1\linewidth]{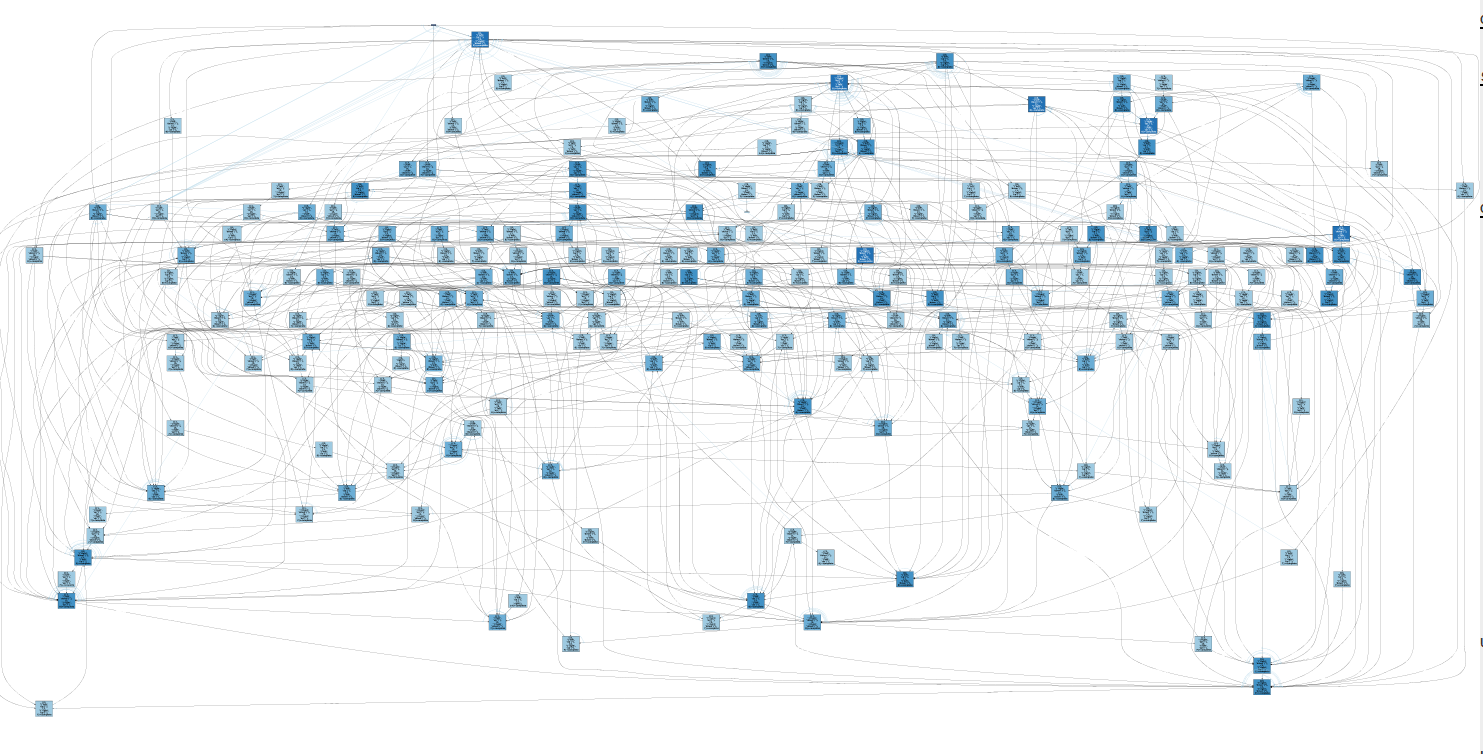}
	\caption{White agent C-net generated by iDHM (Fixed Simulation Depth, Fixed Iteration Times, Minimax Search Depth = 3)}
	\label{fig: white_minimax_3_iDHM}
\end{figure}

\begin{figure}[!h]
	\centering
	\includegraphics[width=1\linewidth]{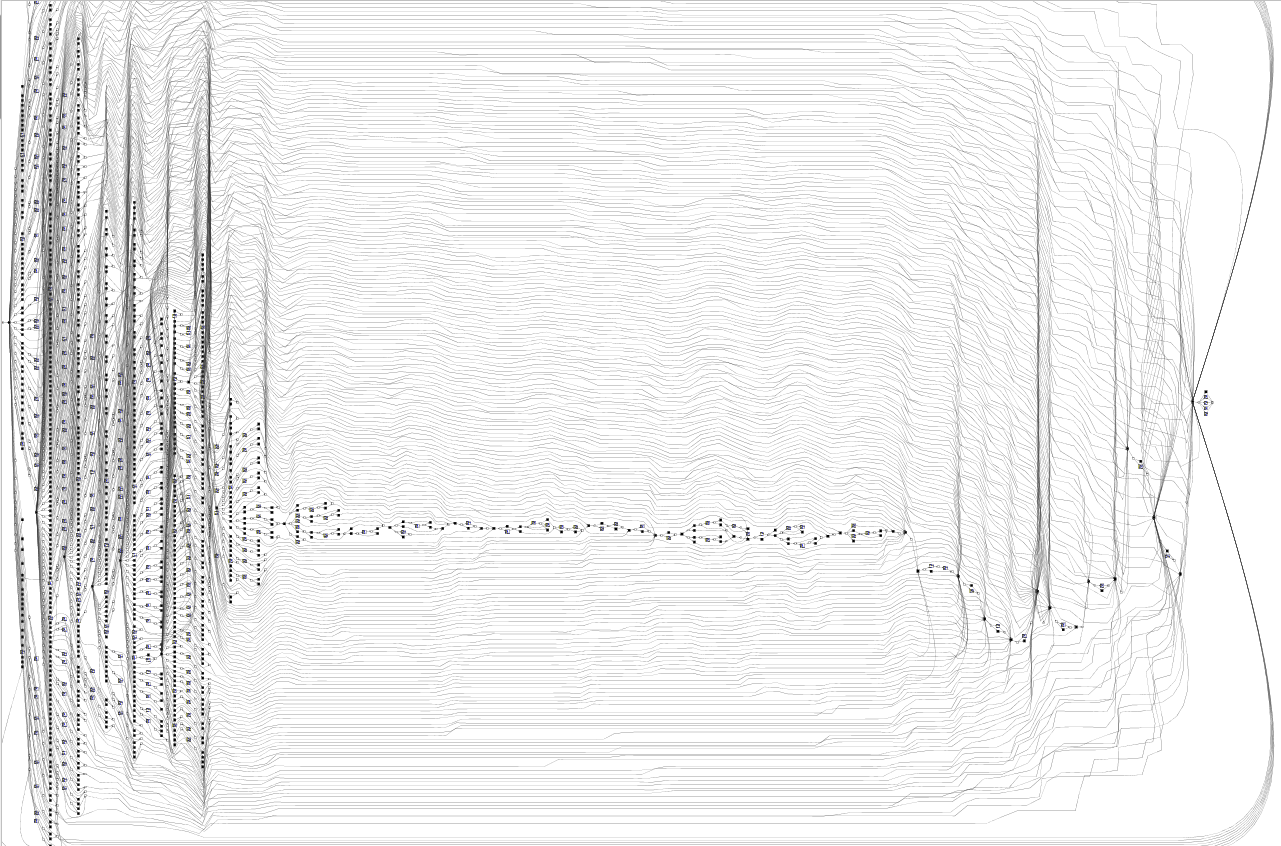}
	\caption{White agent Petri-net generated by inductive miner algorithm (Fixed Simulation Depth, Fixed Iteration Times, Minimax Search Depth = 3)}
	\label{fig: white_minimax_3_inductive}
\end{figure}

\end{document}